\documentclass[lettersize,journal]{IEEEtran}
\usepackage{amsmath,amsfonts}
\usepackage[ruled,linesnumbered]{algorithm2e}
\usepackage{array}
\usepackage{graphicx}
\usepackage{subcaption}
\usepackage{textcomp}
\usepackage{stfloats}
\usepackage{url}
\usepackage{verbatim}

\usepackage{cite}
\usepackage{indentfirst}
\usepackage{multicol}
\usepackage{etoolbox}
\usepackage{relsize}
\usepackage{mathrsfs}
\usepackage{hyperref}
\usepackage{booktabs}
\usepackage{tabularx}
\usepackage{multirow}
\usepackage{colortbl}
\usepackage{xcolor}
\usepackage{amssymb}
\usepackage{bm}
\usepackage{amsthm}
\usepackage{multirow}
\newtheoremstyle{normalfont} % Name
  {3pt}                      % Space above
  {3pt}                      % Space below
  {}                         % Body font (normal font here)
  {}                         % Indent amount
  {\bfseries}                % Theorem head font (bold here)
  {.}                        % Punctuation after theorem head
  { }                        % Space after theorem head
  {}                         % Theorem head spec (empty here)
\theoremstyle{normalfont}
\newtheorem{proposition}{\textbf{Proposition}}
\renewenvironment{proof}{{\bfseries \noindent Proof.}}{\qed}

\def\BibTeX{{\rm B\kern-.05em{\sc i\kern-.025em b}\kern-.08em
    T\kern-.1667em\lower.7ex\hbox{E}\kern-.125emX}}
\usepackage{balance}

\begin{document}

\title{Hybrid Memetic Search for Electric Vehicle Routing with Time Windows, Simultaneous Pickup-Delivery, and Partial Recharges}

% \author{Anonymous authors}
\author{Zubin~Zheng,
        Shengcai~Liu,~\IEEEmembership{Member,~IEEE},
        and Yew-Soon~Ong,~\IEEEmembership{Fellow,~IEEE}

\thanks{
% Zubin Zheng and Shengcai Liu are with the Department of Computer Science and Engineering, Southern University of Science and Technology, Shenzhen, China.
% Email: zhengzb2021@mail.sustech.edu.cn; liusc3@sustech.edu.cn

% Yew-Soon Ong is with the Centre for Frontier AI Research (CFAR), Agency for Science, Technology and Research (A*STAR), Singapore, and is also with the College of Computing and Data Science, Nanyang Technological University (NTU), Singapore.
% Email: asysong@ntu.edu.sg

This work was supported by Guangdong Major Project of Basic and Applied Basic Research under Grant 2023B0303000010, National Natural Science Foundation of China under Grant 62272210, the Distributed Smart Value Chain programme which is funded under the Singapore RIE2025 Manufacturing, Trade and Connectivity (MTC) Industry Alignment Fund-Pre-Positioning (Award No: M23L4a0001), the MTI under its AI Centre of Excellence for Manufacturing (AIMfg) (Award No: W25MCMF014), and the College of Computing and Data Science, Nanyang Technological University.

Zubin Zheng and Shengcai Liu are with the Guangdong Provincial Key Laboratory of Brain-Inspired Intelligent Computation, Department of Computer Science and Engineering, Southern University of Science and Technology, Shenzhen, China. 
Email: zhengzb2021@mail.sustech.edu.cn; liusc3@sustech.edu.cn.
This work was partially conducted during Zubin's internship at CFAR.

Yew-Soon Ong is with the Centre for Frontier AI Research, Institute of High Performance Computing, Agency for Science, Technology and Research, Singapore, and the College of Computing and Data Science, Nanyang Technological University.
Email: asysong@ntu.edu.sg

Corresponding Author: Shengcai Liu.
}

% \thanks{Manuscript created October, 2020; This work was developed by the IEEE Publication Technology Department. This work is distributed under the \LaTeX \ Project Public License (LPPL) ( http://www.latex-project.org/ ) version 1.3. A copy of the LPPL, version 1.3, is included in the base \LaTeX \ documentation of all distributions of \LaTeX \ released 2003/12/01 or later. The opinions expressed here are entirely that of the author. No warranty is expressed or implied. User assumes all risk.}
}

\markboth{Journal of \LaTeX\ Class Files,~Vol.~X, No.~X, MM~YYYY}%
{Hybrid Memetic Search for Electric Vehicle Routing with Time Windows, Simultaneous Pickup-Delivery, and Partial Recharges}

\maketitle

\begin{abstract}
With growing environmental concerns, electric vehicles for logistics have gained significant attention within the computational intelligence community in recent years.
This work addresses an emerging and significant extension of the electric vehicle routing problem (EVRP), namely EVRP with time windows, simultaneous pickup-delivery, and partial recharges (EVRP-TW-SPD), which has widespread real-world applications.
We propose a hybrid memetic algorithm (HMA) for solving EVRP-TW-SPD.
HMA incorporates two novel components: 
a parallel-sequential station insertion (PSSI) procedure for handling partial recharges that can better avoid local optima compared to purely sequential insertion, and a cross-domain neighborhood search (CDNS) that explores solution spaces of both electric and non-electric problem domains simultaneously.
These components can also be easily applied to various EVRP variants.
To bridge the gap between existing benchmarks and real-world scenarios, we introduce a new, large-scale EVRP-TW-SPD benchmark set derived from real-world applications, containing instances with many more customers and charging stations than existing benchmark instances.
Extensive experiments demonstrate the significant performance advantages of HMA over existing algorithms across a wide range of problem instances.
Both the benchmark set and HMA are to be made open-source to facilitate further research in this area.
% With a growing emphasis on environmental issues, electric vehicles in logistics have gained much attention in the last few years.
% This paper addresses the electric vehicle routing problem with time windows and simultaneous pickup and deliveries (EVRP-TW-SPD), incorporating partial recharges at charging stations, which has wide applications in real life.
% We propose a hybrid memetic algorithm~(HMA) to solve this problem. 
% HMA is distinct from existing approaches in two aspects: a general parallel-sequential station insertion mechanism for partial recharges, with a global search capability to evaluate and adjust station insertion, and a cross-domain neighborhood search that leverages existing efficient vehicle routing problem local search methods, expanding the search space beyond the EVRP-TW-SPD neighborhood.
% The application of both components can be easily generalized to other problem variants. 
% Experimental results on public benchmarks demonstrate that HMA outperforms all recent algorithms. 
% Moreover, a comprehensive ablation study is also conducted to show the effectiveness of the novel components integrated in HMA.
% Finally, a new benchmark of large-scale instances derived from a real-world application of JD Logistics is introduced as a challenging dataset for future research.
\end{abstract}


\section{Introduction}
\label{sec: Introduction}
\IEEEPARstart{O}{ver} the past few decades, the vehicle routing problem (VRP) has attracted much attention in the computational intelligence community~\cite{wen2024transfer,FengHZZGTT21,chen2021heuristic}, due to its extensive real-world applications.
As VRP research advances, a new focus has emerged: the electric vehicle routing problem (EVRP)~\cite{mandziuk2018new}.
This area has garnered significant interest in recent years, as electric vehicles (EVs) become increasingly prevalent in various sectors~\cite{qin2021review,zhou2019blockchain,liu2022privacy}, such as public transit, home deliveries, postal services, distribution tasks, and emergency power supply.
While EVs enable zero-emission logistics, operating a fleet of EVs presents additional challenges compared to operating conventional vehicles:
i)~the lower energy density of batteries compared to the fuel of combustion engine vehicles, necessitating EVs to perform detours to recharge batteries;
ii)~the limited number of public charging stations; and 
iii)~the substantial time required for battery charging.
To address these challenges, Erdogan and Miller-Hooks~\cite{erdougan2012green} introduced the first EVRP model that incorporated charging stations, as part of their broader study on the green VRP (G-VRP).
Built on their work, subsequent studies have progressively incorporated more constraints into EVRP models to reflect real-world scenarios~\cite{Kucukoglu2021electric,wang2024deep,ShiZLZ22Memory,Liang2020AnEfficient}.

% However, with the increasing maturity of research in traditional VRP, the rise of electric vehicles (EVs) as a sustainable and emerging mode of transportation has sparked interest in the electric vehicle routing problem (EVRP), positioning it as a new research hotspot in the past decade.
% EVs are now being utilized for a wide range of purposes, including public transit, home deliveries, postal services, and various distribution tasks.
% as an emerging mode of transportation, can be utilized for various needs, including public transit, home deliveries from grocery stores, postal deliveries, and distribution tasks. 
% While EVs enable zero-emission logistics services, operating a fleet of EVs presents several challenges, which hinder their usage on a massive scale~\cite{mandziuk2018new}: 
% i) the~lower energy density of batteries compared to the fuel of combustion engine vehicles, forcing EVs to perform longer detours to recharge batteries; 
% ii) the~limited number of public charging stations; and 
% iii) the~non-negligible charging times. 
% Considering these constraints, Erdogan and Miller-Hooks~\cite{erdougan2012green} introduced the first routing model that incorporates charging stations. 
% Schneider~\textit{et~al.}~\cite{schneider2014electric} extended the model to the Electric Vehicle Routing Problem with Time Windows and Recharge Stations (E-VRPTW).
% With the increasing adoption of EVs and the expansion of charging station infrastructure, EVRP and its variants hold significant potential in industrial applications.
% research general background 

This work focuses on an emerging and significant extension of EVRP, namely the electric vehicle routing problem with time windows, simultaneous pickup-delivery, and partial recharges (EVRP-TW-SPD), introduced by Akbay~\textit{et~al.}~\cite{akbay2022application}.
Compared to the basic EVRP model, EVRP-TW-SPD considers more realistic problem characteristics and constraints, encompassing a wider range of real-world scenarios.
Specifically, this model includes constraints such as limited driving range that requires recharging along the route and strict customer service time windows~\cite{schneider2014electric}.
Additionally, it also integrates reverse logistics and partial recharges.
In reverse logistics~\cite{dethloff2001vehicle,wassan2008reactive}, customers have both pickup and delivery demands, requiring EVs to complete both services in one trip.
Partial recharges~\cite{keskin2016partial,desaulniers2016exact} mean that EVs are not necessarily fully charged each time they visit a charging station, saving time to serve customers with tight time windows.

Despite the extensive real-world applications of EVRP-TW-SPD, current research in this area remains limited~\cite{akbay2022application,akbay2023cmsa,feng2023admm}.
Specifically, existing approaches for solving EVRP-TW-SPD can be categorized into approximation methods~\cite{feng2023admm}
and metaheuristic methods~\cite{akbay2022application,akbay2023cmsa}.
However, neither approach has shown fully satisfactory performance.
Our experiments reveal that even for medium-scale instances with around 100 customers, there is significant room for improvement in solution quality obtained by existing methods.
Furthermore, there is a notable gap between the scale of existing benchmark instances and real-world scenarios.
The largest instance in the publicly available benchmark~\cite{akbay2022application} contains only 100 customers and 21 charging stations.
In contrast, with the rapid growth of urban areas, a real-world EVRP-TW-SPD instance might involve many more customers and stations~\cite{liu2021memetic,wang2023partial}.
For example, application data from JD Logistics in Beijing~\footnote{JD Logistics is one of the largest logistics companies in China.}, a major city in China, shows scenarios with up to 1000 customers and 100 charging stations.
This disparity highlights the need for establishing larger-scale benchmarks to drive research that better addresses practical challenges.

This work aims to address the above research limitations.
Specifically, we propose a novel hybrid memetic algorithm (HMA) for solving EVRP-TW-SPD.
HMA combines individual-based large neighborhood search with population-based memetic search, leveraging their complementary strengths to quickly find high-quality solutions and further improve solution quality.
% The individual-based search offers efficient exploration to quickly find high-quality solutions, while the population-based search further improves solution quality.
Importantly, HMA introduces two key novelties.
First, it incorporates a parallel-sequential station insertion procedure (PSSI) for handling partial recharges.
Compared to the sequential station insertion procedure commonly used in the literature~\cite{keskin2016partial,keskin2018matheuristic,keskin2021simulation}, PSSI can better avoid local optima and construct higher-quality routes with acceptable additional computational cost.
Second, HMA employs a cross-domain neighborhood search (CDNS) as its local search module.
CDNS simultaneously explores two different neighborhoods: those of the non-electric counterpart of EVRP-TW-SPD (i.e., VRP-TW-SPD) and EVRP-TW-SPD itself.
This expanded search space not only leads to higher-quality solutions but also allows for easy reuse of efficient local search procedures from non-electric VRP variants, which are often more extensively studied.
Finally, to address the gap in the scale of benchmark instances, we propose a new benchmark set based on real customer data from JD Logistics in the central area of Beijing.

The main contributions of this work are summarized below.
\begin{enumerate}
\item We propose a novel hybrid memetic algorithm (HMA) for solving EVRP-TW-SPD.
HMA integrates two novel components: PSSI for handling partial recharges and CDNS as the local search module.
We further discuss the potential of applying them to a wide range of EVRP variants beyond EVRP-TW-SPD.
\item We introduce a new, large-scale EVRP-TW-SPD benchmark set derived from real-world applications.
This set, which will be made open-source, contains instances with up to 1000 customers and 100 charging stations, aiming to drive research in the field.
\item Extensive experiments validate the effectiveness of HMA, demonstrating its significant performance advantage over existing methods across a wide range of problem instances. 
\end{enumerate}

The rest of this paper is organized as follows.
Section~\ref{sec: Related Work} presents a literature review.
Section~\ref{sec: Notations and Problem Definition} formally defines the problem.
Section~\ref{sec: Hybrid Algorithm for EVRP-TW-SPD} first describes the novel components of HMA, followed by the hybrid framework. Section~\ref{sec: Computational Study} compares the proposed algorithm against existing methods using the existing benchmark and introduces the new benchmark.
Finally, Section~\ref{sec: Conclusion} concludes the paper.
  
\section{Related Work}
\label{sec: Related Work}
In this section, we first review literature related to two key aspects of EVRP-TW-SPD: reverse logistics and partial recharges, followed by existing methods for the problem.

Reverse logistics, which involves the bi-directional flow of goods in pickup and delivery, is crucial in modern transportation due to its importance in satisfying distinct customer demands and lowering energy consumption. 
According to~\cite{berbeglia2007static}, VRPs involving reverse logistics have often been referred to as pickup and delivery problems (PDPs).
Among PDPs, the VRP with simultaneous pickup-delivery~(VRPSPD) is the most studied variant~\cite{kocc2020review,TangLYY21}, where vehicles deliver goods from a central depot to customers while also picking up items from customers to return to the depot.
VRPSPD was first studied in \cite{min1989multiple} for a book distribution system with 22 customers and two vehicles, demonstrating that combining pickup and delivery operations leads to significant savings in time and distance compared to separate operations.
Since then, numerous studies have been conducted on VRPSPD and its variants.
In EVRP-TW-SPD, to handle simultaneous pickup-delivery, existing techniques from VRPSPD research~\cite{montane2002vehicle,gajpal2010saving,dethloff2001vehicle} can be adapted.
Specifically, in this work we integrate the residual capacity and radial surcharge (RCRS)~\cite{dethloff2001vehicle} construction heuristic into the HMA algorithm.

Compared to full recharges, partial recharges are a more realistic and practical charging scheme in the real world.
It allows for flexible charging time, which can help EVs meet tight time windows and potentially serve more customers.
To handle partial recharges, two key aspects need to be considered: station selection and charge amount calculation.
First, station selection refers to choosing which charging station EVs should use when they need to recharge. 
Most existing approaches for station selection rely on heuristic insertion strategies.
Some approaches, like those in~\cite{mao2020electric,lu2024study}, focus on minimizing the distance from the current customer to the nearest charging station.
Others, such as~\cite{akbay2022application,akbay2023cmsa,keskin2016partial,roberti2016electric}, aim to minimize the increase in total distance caused by inserting a charging station.
Specifically, Keskin~\textit{et al.}~\cite{keskin2016partial} proposed three station insertion heuristics: greedy station insertion, greedy station insertion with comparison, and best station insertion.
In contrast, Wang~\textit{et~al.}~\cite{wang2023partial} took a different approach by minimizing a generalized cost function that includes the objective function, time window, and distance violations.
Second, charge amount calculation involves determining how much to recharge at each charging station.
Several approaches have been proposed.
Felipe~\textit{et~al.}~\cite{felipe2014heuristic} suggested charging only the minimum amount needed to reach the next charging station or depot.
Keskin~\textit{et~al.}~\cite{keskin2016partial} ensured sufficient charge to reach the next station or depot while adjusting for time window constraints.
They also found that partial recharges perform significantly better than full recharges, even with pre-determined charging amounts.
Desaulniers~\textit{et~al.}~\cite{desaulniers2016exact} and Wang~\textit{et~al.}~\cite{wang2023partial} considered both the minimum required charge and available slack time when determining charge amount.
Similar to~\cite{desaulniers2016exact,wang2023partial}, our approach for determining charge amount also considers both the minimal required charge and available slack time, ensuring sufficient battery state while meeting all time constraints.

Currently, research on EVRP-TW-SPD still remains scarce.
Akbay~\textit{et~al.}~\cite{akbay2022application} introduced the problem and formulated it as a mixed integer linear programming (MILP) model.
They proposed a metaheuristic approach called Adapt-CMSA-STD and compared it with CPLEX and other heuristics on instances with up to 100 customers and 21 charging stations.
Their follow-up work~\cite{akbay2023cmsa} presented Adapt-CMSA-SETCOV, which used vehicle-route representations and a set-covering-based MILP model and showed performance improvement over Adapt-CMSA-STD.
Feng~\textit{et~al.}~\cite{feng2023admm} developed an alternating direction multiplier method for EVRP-TW-SPD with full recharges, testing it on small-scale instances with up to 15 customers and 6 charging stations.
\section{Notations and Problem Definition}
\label{sec: Notations and Problem Definition}
Formally, the EVRP-TW-SPD is defined on a complete directed graph \( \bm{G} = (\bm{V}', \bm{E}) \), where \( \bm{V}' = \{0\} \cup \bm{V} \cup \bm{F} = \{0, 1, 2, \ldots, M + P\} \) is the node set.
Here, \( 0 \) represents the depot, \( \bm{V} = \{1, 2, \ldots, M\} \) is a set of \( M \) customers, and \( \bm{F} = \{M + 1, \ldots, M + P\} \) is a set of \( P \) charging stations allowing multiple visits.
The arc set \( \bm{E} = \{ (i, j) \mid i, j \in \bm{V}', i \neq j \} \) is defined between each pair of nodes.
Each node \( i \in \bm{V}' \) has five attributes: delivery demand \( u_i \), pickup demand \( v_i \), time window \([e_i, l_i]\), and service time \( s_i \).
For a customer \( i \), \( u_i \) is the quantity of goods delivered from the depot, and \( v_i \) is the amount of goods picked up to be returned to the depot. 
The time window \([e_i, l_i]\) defines the earliest and latest time for customer service, with arrivals before \( e_i \) resulting in waiting and after \( l_i \) being infeasible.
\( s_i \) is the unloading/loading time at customer \( i \).
For the depot \( 0 \), \( e_0 \) and \( l_0 \) are the earliest departure and latest return time for EVs, respectively, with \( u_0 = v_0 = s_0 = 0 \).
For a charging station \( i \in \bm{F} \), \( u_i = v_i = s_i = 0 \), and \( e_i = e_0 \), \( l_i = l_0 \).
Each arc \( (i, j) \in \bm{E} \) is associated with a travel distance \( d_{ij} \) and a travel time \( t_{ij} \).

A homogeneous fleet of EVs, each with identical loading capacity \(C\), battery capacity \(Q\), and dispatching cost \(\mu_1\), is initially located at the depot.
The EVs depart from the depot, serve customers, and finally return to the depot.
When an EV visits a charging station, its battery is recharged at a constant rate of \(g > 0\), i.e., it takes time \(g\) to charge one unit of battery energy.
Conversely, the constant \(h\) represents the EV's battery energy consumption rate per unit of travel distance for the EV.

A solution $\bm{S}$ to EVRP-TW-SPD is represented by a set of vehicle routes,  \(\bm{S} = \{\bm{R}_1, \bm{R}_2, \ldots, \bm{R}_K\}\), where \(K\) is the number of vehicles used.
Each route \(\bm{R}_i\) consists of a sequence of nodes that the EV visits, \(\bm{R}_i = (n_{i,1}, n_{i,2}, \ldots, n_{i, L_i})\), where \(n_{i,j}\) is the \(j\)-th node visited in \(\bm{R}_i\), and \(L_i\) is the length of \(\bm{R}_i\).
For brevity, we temporarily omit the subscript \(i\) in \(\bm{R}_i\) in the following, i.e., \(\bm{R} = (n_1, n_2, \ldots, n_L)\).
The total travel distance of \(\bm{R}\), denoted as \(TD(\bm{R})\), is:
\begin{equation}
\begin{aligned}
TD(\bm{R}) &= \sum_{j=1}^{L-1} d_{n_j n_{j+1}}.
\end{aligned}
\label{eq:1}
\end{equation}
The battery energy state of the EV on arrival at and departure from \(n_j\), denoted as \(y_{n_j}\) and \(Y_{n_j}\), respectively, can be computed recursively as follows:
\begin{equation}
\begin{aligned}
y_{n_j} &= Y_{n_{j-1}} - h \cdot d_{n_{j-1}n_j}, \ j > 1, \\
Y_{n_j} &= 
\begin{cases}
Q, & \ j = 1, \\
y_{n_j}, & \ j > 1, \ n_j \notin \bm{F}, \\
y_{n_j} + q_{n_j}, & \ j > 1, \ n_j \in \bm{F},
\end{cases}
\end{aligned}
\label{eq:2}
\end{equation}
where \(q_{n_j}\) is the charge amount at charging station \(n_j \in \bm{F}\).
An EV departs from the depot with a full battery.
% but it is not necessary for the EV to return to the depot with a completely depleted battery if it has been recharged at least once along its 
The arrival and departure time at \(n_j\), denoted as \(a_{n_j}\) and \(b_{n_j}\), respectively, can be computed recursively as follows:
\begin{equation}
\begin{aligned}
a_{n_j} &= b_{n_{j-1}} + t_{n_{j-1}n_j}, \ j > 1, \\
b_{n_j} &= \begin{cases} 
e_0, & \ j = 1, \\
\max \{ a_{n_j}, e_{n_j} \} + s_{n_j}, & \ j > 1,  \ n_j \notin \bm{F},   \\ 
a_{n_j} + g \cdot q_{n_j},&  \ j > 1, \  n_j \in \bm{F}. \\ 
\end{cases}  
\end{aligned}
\label{eq:3} 
\end{equation}
The EV load on arrival at \(n_j\), denoted as \(load_{n_j}\), is:
\begin{equation}
\begin{aligned}
% load_{n}_1 &= \sum_{j=1}^{L}u_{n_j}, \\
load_{n_j} &= \begin{cases}
\sum_{j=1}^{L}u_{n_j}, & \ j = 1, \\
load_{n_{j-1}} - u_{n_{j-1}} + v_{n_{j-1}}, & \ j > 1.
\end{cases} 
\end{aligned}
\label{eq:4} 
\end{equation}

The total cost of \(\bm{S}\), denoted as \(TC(\bm{S})\), consists of two parts: the dispatching cost, \(\mu_1 \cdot K\), and the transportation cost, which is the total travel distance of \(\bm{S}\) multiplied by the cost per unit of travel distance \(\mu_2\).
The objective is to find a solution with the minimum \(TC\), as presented in Eq.~\eqref{eq:5}:
\begin{subequations}
\begin{align}
    &\min_{\bm{S}} TC(\bm{S}) = \mu_1 \cdot K + \mu_2 \cdot \sum_{i=1}^{K} TD(\bm{R}_i) \label{eq:5} \\
    \text{s.t.} 
    & \quad n_{i,1} = n_{i,L_i} = 0, \ \forall i \in \{1, \dots, K\}, \label{eq:6} \\
    & \quad \sum_{i=1}^{K} \sum_{j=2}^{L_i - 1} \mathbb{I}[n_{i,j} = z] = 1, \ \forall z \in \{1, \dots, M\}, \label{eq:7} \\
    &\quad 0 \leq load_{n_{i,j}} \leq C, \ \forall i \in \{1, \dots, K\}, \ \forall j \in \{1, \dots, L_i\}, \label{eq:8} \\
    &\quad a_{n_{i,j}} \leq l_{n_{i,j}}, \ \forall i \in \{1, \dots, K\}, \ \forall j \in \{2, \dots, L_i\}, \label{eq:9} \\
    &\quad b_{n_{i,1}} \geq e_0 \ \text{and} \ a_{n_{i,L_i}} \leq l_0, \ \forall i \in \{1, \dots, K\}, \label{eq:10} \\
    &\quad 0 \leq y_{n_{i,j}} \leq Y_{n_{i,j}} \leq Q, \ \forall i \in \{1, \dots, K\}, \ \forall j \in \{1, \dots, L_i\}.   \label{eq:11}
\end{align}
\end{subequations}
% \begin{subequations}
% \begin{align}
%     &\min_{\bm{S}} TC(\bm{S}) = \mu_1 \cdot K + \mu_2 \cdot \sum_{i=1}^{K} TD(\bm{R}_i) \label{eq:5} \\
%     \text{s.t.} 
%     & \quad n_{i,1} = n_{i,L_i} = 0, \ 1 \leq i \leq K, \label{eq:6} \\
%     & \quad \sum_{i=1}^{K} \sum_{j=2}^{L_i - 1} \mathbb{I}[n_{i,j} = z] = 1, \ 1 \leq z \leq M, \label{eq:7} \\
%     &\quad 0 \leq load_{n_{i,j}} \leq C, \ 1 \leq i \leq K, \ 1 \leq j \leq L_i, \label{eq:8} \\
%     &\quad a_{n_{i,j}} \leq l_{n_{i,j}}, \ 1 \leq i \leq K, \ 2 \leq j \leq L_i, \label{eq:9} \\
%     &\quad b_{n_{i,1}} \geq e_0 \ \text{and} \ a_{n_{i,L_i}} \leq l_0, \ 1 \leq i \leq K, \label{eq:10} \\
%     &\quad 0 \leq y_{n_{i,j}} \leq Y_{n_{i,j}} \leq Q, \ 1 \leq i \leq K, \ 1 \leq j \leq L_i.   \label{eq:11}
% \end{align}
% \end{subequations}
Constraint \eqref{eq:6} ensures that each route starts from and returns to the depot.
Constraint \eqref{eq:7} ensures that each customer is served exactly once (note \( \mathbb{I}[\cdot] \) is the indicator function).
Constraint \eqref{eq:8} ensures that the delivery and pickup demands of customers are met simultaneously while EVs do not exceed their loading capacity \(C\).
Constraint \eqref{eq:9} requires EVs to visit each node within its corresponding time window. Constraint \eqref{eq:10} ensures that EVs depart only after the start of the depot's time window and return before the end of its time window.
Finally, constraint \eqref{eq:11} requires EVs to maintain a non-negative battery state at all nodes; when serving customers or visiting charging stations, the battery level will not decrease, and it will not exceed battery capacity \(Q\) when leaving any node.

In this work, unless otherwise specified, we evaluate solutions based on their cost (as their fitness) defined in Eq.~(\ref{eq:5}), where a lower cost indicates a better solution.

\section{Hybrid Memetic Algorithm for EVRP-TW-SPD}
\label{sec: Hybrid Algorithm for EVRP-TW-SPD}
In this section, we first introduce the two novel components integrated into HMA: PSSI for handling partial recharges and CDNS as the local search module. 
Then, we describe the overall framework of HMA.
% Finally, the decomposition to further efficiently handle large-scale EVRP-TW-SPD are shown.
% In this section, we first elaborate on two important components of HMA, parallel-sequential station insertion and Cross-Domain Neighborhood Search, and then describe the hybrid search framework.
\subsection{Parallel-Sequential Station Insertion (PSSI)}
\label{subsec: Parallel-Sequential Station Insertion}
% 1. general application
% 2. Problem? 满足什么条件 
% 3. example?

% 2. Problem? 满足什么条件: 
% 不同条件：
% (1) PSI: at most one charging station may be visited between nodes; SSI: allows the EV to visit multiple charging stations between nodes in the route (PSI有全局搜索能力，但是适用范围有约束，比较适合较短的route； SSI适用范围广泛，但是解构建单一)
% 相同条件：
% (1) E-VRP with time windows
% (2) a partial linear recharging policy
% (3) the network of charging stations is homogeneous

To transform electricity-infeasible routes~(i.e., routes that do not meet constraint~\eqref{eq:11}) into high-quality feasible routes, we propose a parallel-sequential station insertion (PSSI) procedure.
Compared to existing sequential insertion procedures~\cite{keskin2016partial,keskin2018matheuristic,keskin2021simulation} that typically lead to suboptimal routes, PSSI can better avoid local optima and generate higher-quality routes with acceptable additional computational cost.
As shown in Algorithm~\ref{alg: Parallel-Sequential Station Insertion}, PSSI employs two distinct strategies: parallel station insertion (PSI) and sequential station insertion (SSI).
PSI (lines~\ref{algline: PSI SR}--\ref{algline: PSI Select Best}) employs a genetic algorithm to search for the best station insertions between all consecutive customer pairs simultaneously, determining both the necessity of insertions and the inserted stations.
% uses a binary encoding to simultaneously represent all the possible insertions of charging stations between consecutive customers in a 
% Then, it employs a genetic algorithm to search for high-quality routes, incorporating local adjustments to address any remaining electricity infeasibility.
SSI (lines~\ref{algline: SSI temp}--\ref{algline: SSI Station Refinement}), on the other hand, inserts charging stations sequentially along the route.
% It considers the insertion of multiple stations between consecutive customers when necessary.
Finally, PSSI selects the better route between PSI and SSI as the final outcome (line~\ref{algline: best in PSI and SSI}).
By combining PSI and SSI, PSSI enables a more comprehensive exploration of the route space, potentially yielding higher-quality routes compared to purely sequential procedures.

As aforementioned in Section~\ref{sec: Related Work}, handling partial recharges involves two key aspects: station selection and charge amount calculation. 
In terms of station selection, PSI is designed for scenarios where at most one charging station is needed between any two consecutive nodes, while SSI allows multiple charging stations to be inserted between two consecutive nodes.
Regarding charge amount calculation, both PSI and SSI use the same method.
% to calculate the charge amount at each selected station.
Below, we detail PSI and SSI based on these two aspects.

% As aforementioned in Section~\ref{sec: Related Work}, station selection and charge amount calculation are two important aspects of handling partial recharges. PSI addresses situations where at most one charging station may be visited between nodes, whereas SSI allows the EV to visit multiple charging stations between two consecutive nodes. This distinction arises because PSI and SSI differ in station selection, although they share the same charge amount calculation. Below, we introduce our parallel-sequential station insertion based on these two aspects.
\begin{algorithm}[tbp]
    \small
    \caption{PSSI}
    \label{alg: Parallel-Sequential Station Insertion}
    \LinesNumbered
    \KwIn{an electricity-infeasible route: $\bm{R}$}
    \KwOut{the best-found feasible route: $\bm{R}^*$}

    \tcc{Parallel Station Insertion~(PSI)}
    $\bar{\bm{R}} \gets$ remove all stations in $\bm{R}$\; \label{algline: PSI SR}
    $\{(\bm{x}_1, \bm{R}_1), \ldots, (\bm{x}_{\alpha L}, \bm{R}_{\alpha L}) \} \gets $ Initialization($\bar{\bm{R}},\alpha$)\;  \label{algline: PSI Init}
    \While{max generation count $B$ not reached}{  \label{algline: PSI Iter}
        \For{$i \gets 1$ \KwTo $\alpha L$}{ \label{algline: PSI FOR}
            $\bm{x}_{\text{p}_1},\bm{x}_{\text{p}_2} \gets \text{Random\_Select}(\bm{x}_1, \bm{x}_2, \ldots, \bm{x}_{\alpha L})$\; \label{algline: PSI Random Select}
            $\bm{x}_{\text{child}} \gets \bm{x}_{\text{p}_1} \oplus \bm{x}_{\text{p}_2}$\; \label{algline: PSI Crossover}
            $\bm{x}_{\text{child}} \gets \text{Mutation}(\bm{x}_{\text{child}})$\; \label{algline: PSI Mutation}
            $\bm{R}_{\text{child}} \gets$ insert stations into $\bar{\bm{R}}$ according to $\bm{x}_{\text{child}}$\; \label{algline: PSI Route Evaluation}
            \lIf{$\bm{R}_{\mathrm{child}}$ is better than $\bm{R}_i$}{$(\bm{x}_i, \bm{R}_i) \gets (\bm{x}_{\text{child}}, \bm{R}_{\text{child}})$} \label{algline: PSI Update}
        }  \label{algline: PSI ENDFOR}
    }  \label{algline: PSI EndIter}
    $\bm{R}_{\text{PSI}} \gets \text{Select\_Best}(\bm{R}_1, \ldots, \bm{R}_{\alpha L})$\; \label{algline: PSI Select Best}

    \tcc{Sequential Station Insertion~(SSI)}
    $\bm{R}' \gets \bm{R}$\; \label{algline: SSI temp}
    \While{$\bm{R}'$ is electricity-infeasible}{  \label{algline: SSI While}
        $n_{\text{right}} \gets$ the first node with negative battery state in $\bm{R}'$\; \label{algline: SSI r}
        $n_{\text{left}} \gets$ the last station or depot visted before $n_{\text{right}}$ in $\bm{R}'$\; \label{algline: SSI l}
        $\bm{R}' \gets$ perform the best-possible station insertion in the path from $n_{\text{left}}$ to $n_{\text{right}}$\; \label{algline: SSI Best Station Insertion}
    } \label{algline: SSI Endwhile}
    $\bm{R}_{\text{SSI}} \gets$ improve consecutive stations in $\bm{R}'$\; \label{algline: SSI Station Refinement}

    $\bm{R}^* \gets \text{Select\_Best}(\bm{R}_{\text{PSI}}, \bm{R}_{\text{SSI}})$\; \label{algline: best in PSI and SSI}

    \textbf{return} $\bm{R}^*$
\end{algorithm}
\subsubsection{Station Selection}
For a given problem instance \( \bm{G} = (\bm{V}', \bm{E}) \), we preprocess the charging stations to rank them for insertion between each pair of nodes.
The ranking metric is the extra cost induced by insertion, formally defined as $d_{ik}+d_{kj}-d_{ij}$, where $k$ is a potential charging station to be inserted between nodes $i$ and $j$.
The ranking is in ascending order of this extra cost.
To balance efficiency and solution quality, only the top $sr \cdot |\bm{F}|$ ranked charging stations are considered for insertion in both PSI and SSI, where $sr \in (0, 1]$ is the selection range parameter.

PSI utilizes a genetic algorithm that consists of two main stages: initialization and evolution.
In the initialization stage~(lines~\ref{algline: PSI SR}--\ref{algline: PSI Init}), all charging stations are first removed from the input electricity-infeasible route $\bm{R}$, resulting in a fixed route \(\bar{\bm{R}} = (c_0, c_1, c_2, \ldots, c_{L-1}, c_L)\), where $c_0 = c_L = 0$ represent the depot, and $c_1, c_2, \ldots, c_{L-1}$ are the $L-1$ customers. 
We define a binary decision variable \(X_{j, j+1}\) to indicate whether a charging station is inserted between \(c_j\) and \(c_{j+1}\) (1 for insertion, 0 for no insertion).
Thus, the station insertion information is represented as a binary sequence \(\bm{x} = (X_{0,1}, X_{1,2}, \ldots, X_{L-1, L})\).
An initial population of $\alpha L$ binary sequences (i.e., the population size is proportional to $L$, with 
$\alpha$ representing the proportionality parameter) is first generated, where each $X_{j, j+1}$ is randomly set to 1 or 0 with equal probability~(line~\ref{algline: PSI Init}).
For each sequence, the binary encoding determines whether to insert a station between two consecutive nodes, and we always attempt to insert the highest-ranked station when insertion is indicated.
% we insert charging stations according to the binary encoding, always attempting to insert the highest-priority station first.
If this insertion violates constraints, we try the next highest-ranked station until a feasible insertion is found or all stations within the selection range have been attempted.
Since the initial binary value is assigned randomly, it may fail to find a feasible insertion. When this happens, the sequence is regenerated iteratively until a feasible one is found.
This process results in a set of feasible routes $\bm{R}_1, \bm{R}_2, \ldots, \bm{R}_{\alpha L}$.

In the evolution stage~(lines~\ref{algline: PSI Iter}--\ref{algline: PSI Select Best}), the population evolves over $B$ generations.
In each generation, two parent sequences are first randomly selected~(line~\ref{algline: PSI Random Select}).
A child sequence $\bm{x}_{\text{child}}$ is generated using bitwise XOR crossover (denoted by $\oplus$), inheriting unique insertion positions from each parent~(line~\ref{algline: PSI Crossover}).
Then, mutation is applied to $\bm{x}_{\text{child}}$: each bit has a low chance of flipping (i.e., 2\%), followed by a high chance of flipping from 1 to 0 (i.e., 20\%) to encourage fewer station insertions~(line~\ref{algline: PSI Mutation}).
The corresponding route $\bm{R}_{\text{child}}$ is evaluated and directly discarded if it is infeasible~(line~\ref{algline: PSI Route Evaluation}). 
Finally, a sequence in the population is replaced with $\bm{x}_{\text{child}}$ if $\bm{R}_{\text{child}}$ is with higher quality~(line~\ref{algline: PSI Update}). 
The best route $\bm{R}_{\text{PSI}}$ among the final population is selected as the outcome of PSI (line~\ref{algline: PSI Select Best}).

Unlike PSI, SSI progressively inserts stations along the route (lines~\ref{algline: SSI While}--\ref{algline: SSI Endwhile}).
Specifically, for the first node $n_{\text{right}}$ in the route with a negative battery level, it first finds the last charging station or depot visited before this node, denoted as $n_{\text{left}}$.
Then, it evaluates all potential station insertions in the path from $n_{\text{left}}$ to $n_{\text{right}}$. 
Finally, SSI selects the best feasible insertion to perform, i.e., the one that increases the route distance the least (lines~\ref{algline: SSI r}--\ref{algline: SSI Best Station Insertion}).
However, this greedy procedure can be short-sighted, particularly when multiple stations are inserted between two consecutive nodes.
Among these multiple stations, the station inserted earlier may no longer be the best choice.
To address this limitation, the SSI procedure includes an additional refinement step (line~\ref{algline: SSI Station Refinement}).
That is, for consecutive stations in the route, we attempt to replace them with stations based on the preprocessed rankings again, starting from the highest-ranked to lower-ranked stations.
If an improvement is found, the replacement is made.
The final outcome of the SSI procedure is $\bm{R}_{\text{SSI}}$.

\subsubsection{Charge Amount Calculation}
We propose a new procedure tailored to EVRP-TW-SPD for calculating the partial recharge amount, building upon the work of~\cite{desaulniers2016exact,wang2023partial}.
Our approach considers both the minimal required charge amount and the available slack time.
This ensures that the battery state is sufficient to reach the next charging station (or the depot) while satisfying time window constraints.
Moreover, it aims to minimize waiting time by utilizing any available slack time for additional charging.
% Our approach aims to minimize waiting time by utilizing any available slack time for additional charging.

We define a path \(\bm{P}_i = (f_{i}, c_{i,1}, c_{i,2}, \ldots, c_{i,m_i}, f_{i+1})\) in route \(\bm{R}\), where \(f_{i}\) and \(f_{i+1}\) are two successive charging stations (or the depot) in the route, separated by \( m_i \) customers \( c_{i,j}\ (j=1,...,m_i) \).
The charge amount at \(f_i\) consists of two parts: the minimal charge \( q_{f_i, 0}\) and the additional charge \( q_{f_i, 1}\).
Specifically, $q_{f_i, 0}$ is calculated as:
\begin{equation}
\begin{aligned}
q_{f_i, 0} & = \max \Bigg\{0, h \cdot TD(\bm{P}_i) - y_{f_i} \Bigg\},
\end{aligned}
\label{eq:12}
\end{equation}
where \(h\) is the energy consumption rate and \(TD(\bm{P}_i)\) is the total distance of path \(\bm{P}_i\).

To compute the additional charge \( q_{f_i, 1}\),  we define the following variables.
Let $a'_{c_{i,j}}$, $b'_{c_{i,j}}$ be the updated arrival and departure time at $c_{i,j}$ (distinguished from $a_{c_{i,j}}$, $b_{c_{i,j}}$ in Section~\ref{sec: Notations and Problem Definition}), $\delta_{i,j}$ be the slack time that can be used to charge due to waiting before starting service at $c_{i,j}$, $\tau_{i,j}$ be the potential available time that can be used to charge when leaving $c_{i,j}$ while satisfying all previous time window constraints.
Without loss of generality, $f_{i}$ is regarded as $c_{i,0}$ for notation convenience.
Hence, \( q_{f_i, 1}\) can be computed as:
\begin{equation}
\begin{aligned}
q_{f_i, 1} =  \frac{1}{g} \sum_{j=1}^{m_i} \delta_{i,j},
\end{aligned}
\label{eq:13}
\end{equation}
where \(g\) is the charging rate and $\delta_{i,j}$ can be computed recursively as follows:
\begin{equation}
\begin{aligned}
a'_{c_{i,j}} & = b'_{c_{i,j-1}} + t_{c_{i,j-1}c_{i,j}}, \ j > 0,\\
b'_{c_{i,j}} & = \begin{cases} 
a_{f_i} + g \cdot q_{f_i, 0}, & \ j = 0, \\
\max\{e_{c_{i,j}}, a'_{c_{i,j}} + \delta_{i,j}\} + s_{c_{i,j}},  & \ j > 0, \\
\end{cases} \\
\delta_{i,j}  & = \min \{\tau_{i,j-1},  \max\{0, e_{c_{i,j}}-a'_{c_{i,j}}\}\}, \ j > 0,\\
\tau_{i,j} & = \begin{cases}  
\infty, & \ j = 0,\\
\min \{\tau_{i,j-1},  l_{c_{i,j}}-a'_{c_{i,j}}\} - \delta_{i,j} , & \ j > 0.\\
\end{cases} 
\end{aligned}
\label{eq:14} 
\end{equation}
When \(\tau_{i,j}\) is \(0\), the recursive equations can terminate prematurely. 
Upon reaching $c_{i,m_i}$ of \(\bm{P}_i\), if no slack time is available, even if $\tau_{i,m_i} > 0 $, no more charge time is needed. 
Thus, the additional charge \( q_{f_i, 1}\) does not affect the arrival time at the next station $f_{i+1}$.

Finally, the total charge amount \(q_{f_i}\) that does not exceed $Q-y_{f_i}$ when visiting $f_{i}$ is computed as:
\begin{equation}
\begin{aligned}
q_{f_i} &=\min\{ q_{f_i, 0} + q_{f_i, 1}, Q-y_{f_i}\}.\\
\end{aligned}
\label{eq:15} 
\end{equation}

\subsubsection{Discussion}
Compared to existing purely sequential insertion procedures~\cite{keskin2016partial,keskin2018matheuristic,keskin2021simulation},
PSSI can typically obtain higher-quality feasible routes with reasonable additional computational cost, as demonstrated in the experiments (see Section \ref{subsec: Effectiveness of Each Component in HMA}).
The applicability of PSSI extends beyond EVRP-TW-SPD to various EVRP variants sharing certain features:
i) partial recharges, ii) customers with time windows, and iii) a homogeneous network of charging stations.
For instance, PSSI could be applied to scenarios involving multiple depots, heterogeneous vehicle fleets, or even dynamic routing where customer demands or traffic conditions change in real-time.

\subsection{Cross-Domain Neighborhood Search (CDNS)}
\label{sec:CDNS}
We propose cross-domain neighborhood search (CDNS), a novel local search procedure that explores two distinct solution spaces simultaneously: the solution space of EVRP-TW-SPD and the solution space of its non-electric counterpart VRP-TW-SPD~\cite{liu2021memetic}, where some solutions may be infeasible for EVRP-TW-SPD.
The non-electric counterpart, VRP-TW-SPD, only excludes the electricity constraint~\eqref{eq:11} and, as a result, does not impose any vehicle distance limit.
This cross-domain exploration offers two key advantages.
First, by expanding beyond the feasible solution space of EVRP-TW-SPD, CDNS potentially discovers higher-quality solutions that might be overlooked by conventional single-domain local search.
Second, exploring the solution space of VRP-TW-SPD enables the efficient reuse of well-established local search procedures for VRP-TW-SPD.
We will further discuss the broader implications of this cross-domain approach beyond EVRP-TW-SPD at the end of Section~\ref{sec:CDNS}.

The core idea of CDNS is to exploit the relationship between feasible solutions of EVRP-TW-SPD and VRP-TW-SPD~\cite{liu2021memetic}, as stated below.
\begin{proposition}
\label{proposition: VRP-TW-SPD & EVRP-TW-SPD}
Given an EVRP-TW-SPD instance, removing all charging stations from a feasible EVRP-TW-SPD solution always results in a feasible VRP-TW-SPD solution.
\end{proposition}
The proof can be found in Appendix~\ref{app:proof_pro_1}.
Conversely, to transform a feasible VRP-TW-SPD solution to a feasible EVRP-TW-SPD solution, we can use the aforementioned PSSI procedure.

\begin{algorithm}[tbp]
    \small
    \caption{CDNS}
    \label{alg: Cross-Domain Neighborhood Search}
    \LinesNumbered
    \KwIn{a feasible solution: $\bm{S}$}
    \KwOut{a local optimum: $\bm{S}$}
    \tcc{Aggressive Local Search~(ALS)}
    $\bm{S}' \gets \bm{S}$\; \label{algline: CDNS x copy}
    $\bar{\bm{S}} \gets$ remove all stations in $\bm{S}'$\; \label{algline: CDNS Station Remove}
    \Repeat{\rm $\bar{\bm{S}}$ is not improved or $\bm{S}'$ is electricity-infeasible}{ \label{algline: CDNS REPEAT1}
        $\bar{\bm{S}} \gets $ apply an existing local search procedure for VRP-TW-SPD to find the best-improving solution in the neighborhood of $\bar{\bm{S}}$\;
        \label{algline: CDNS Move A}
        \For{\rm each route $\bar{\bm{R}} \in \bar{\bm{S}}$}
        { \label{algline: CDNS FORALL}
            $\bm{R} \gets \text{PSSI}(\bar{\bm{R}})$\;
            replace the corresponding route in $\bm{S}'$ with $\bm{R}$\; \label{algline: CDNS Update}
        } \label{algline: CDNS ENDFOR1}
        %$\bm{S} \gets \text{Select\_Best}(\bm{S}', \bm{S})$\; 
        \lIf{$\bm{S}'$ is better than $\bm{S}$}{$\bm{S} \gets \bm{S}'$} \label{algline: CDNS Select Best}
    } \label{algline: CDNS UNTIL1}

    \tcc{Conservative Local Search~(CLS)}
    \Repeat{\rm $\bm{S}$ is not improved}{ \label{algline: CDNS REPEAT2}
        $\bm{S} \gets$ find the best-improving solution in the EVRP-TW-SPD neighborhood of $\bm{S}$\; \label{algline: CDNS Move C}
    } \label{algline: CDNS UNTIL2}
    \Return{$\bm{S}$}
\end{algorithm}

Algorithm~\ref{alg: Cross-Domain Neighborhood Search} outlines the framework of CDNS.
It consists of two sub-procedures.
The first, called aggressive local search (ALS), explores the solution space of VRP-TW-SPD.
As its name suggests, it temporarily ignores the electricity constraint~\eqref{eq:11} and repeatedly applies existing local search procedures for VRP-TW-SPD to find the solution with maximum improvement (i.e., the best-improving solution) in the neighborhood of the current solution (lines \ref{algline: CDNS x copy}--\ref{algline: CDNS UNTIL1}).
When ALS can no longer improve the solution, CDNS switches to conservative local search (CLS) which searches for improvements within the solution space of EVRP-TW-SPD~(lines~\ref{algline: CDNS REPEAT2}--\ref{algline: CDNS UNTIL2}).
Below we detail these two sub-procedures.

\subsubsection{Aggressive Local Search (ALS)}
When a feasible solution $\bm{S}$ to EVRP-TW-SPD enters ALS (lines~\ref{algline: CDNS x copy}--\ref{algline: CDNS UNTIL1}), the first step is to remove all charging stations in $\bm{S}$, resulting in a feasible solution $\bar{\bm{S}}$ to VRP-TW-SPD~(line~\ref{algline: CDNS Station Remove}).
Then, an existing local search procedure for VRP-TW-SPD is repeatedly applied to find the best-improving solution in the neighborhood of $\bar{\bm{S}}$ (line~\ref{algline: CDNS Move A}).
Here, we employ the local search procedure of MATE~\cite{liu2021memetic}, one of the state-of-the-art algorithms for solving VRP-TW-SPD.
This procedure involves multiple move operators and constant-time move evaluation (see the paper of MATE for details).
Then, for each found best-improving solution, PSSI is first applied to transform it into a feasible EVRP-TW-SPD solution (lines~\ref{algline: CDNS FORALL}--\ref{algline: CDNS ENDFOR1}).
If this transformation is not possible, then ALS would terminate (i.e., the second termination condition in line~\ref{algline: CDNS UNTIL1}) since further exploration of this solution would not lead to a feasible electric version.
If successful, the resulting feasible solution would replace the current solution $\bm{S}$ when it has better quality (line~\ref{algline: CDNS Select Best}).
ALS would terminate when no further improvement is possible (i.e., the first termination condition in line~\ref{algline: CDNS UNTIL1}).

\subsubsection{Conservative Local Search (CLS)}
Unlike ALS, CLS performs local search directly in the solution space of EVRP-TW-SPD.
It repeatedly identifies the best-improving solution in the current solution's neighborhood until no further improvement is possible~(lines~\ref{algline: CDNS REPEAT2}--\ref{algline: CDNS UNTIL2}).
The neighborhood is defined by the following five move operators. 
Note that all solutions generated by applying any of the five operators to $\bm{S}$ are evaluated, with the electricity constraint~\eqref{eq:11} checked to ensure feasibility.
\begin{itemize}
    \item \textit{2-opt}: it inverts a subsequence of two consecutive customers in a route.
    \item \textit{2-opt*}: it removes two arcs from two different routes, splitting each into two parts, then reconnects the first part of one route with the second part of the other, and vice versa.
    \item \textit{or-opt}: it removes one or two consecutive customers from a route and reinserts them elsewhere in the same or a different route. 
    \item \textit{swap}: it exchanges two non-overlapping subsequences of one or two consecutive customers, either within the same route or across different routes.
    \item \textit{relocate}: it moves a customer from its current position to a new position, either within the same route or to a different route.
\end{itemize}
% Each move operator checks the electricity constraint~\eqref{eq:11} at all possible move positions on each route $\bm{R}$, which takes $O(|\bm{R}|)$ time complexity and results in a relatively slower search.

% Then $\bar{\bm{S}}$ performs the best-improvement move in the VRPSPDTW neighborhood using move operators (\textit{2-opt},~\textit{2-opt*},~\textit{or-opt} and~\textit{swap}) and the efficient $O(1)$ time complexity move evaluation both from MATE. 
% The \textit{2-opt} operator inverts a subsequence of two consecutive customers in a route. 
% The \textit{2-opt*} operator removes two arcs from different routes, splitting each into two parts, then reconnects the first part of one route with the second part of the other, and vice versa. 
% The \textit{or-opt} operator removes one or two consecutive customers from a route and reinserts them elsewhere in the same or a different route. 
% The \textit{swap} operator exchanges two non-overlapping subsequences of one or two consecutive customers, either within the same route or across different routes.

% Although ALS efficiently explores a larger solution space, it may terminate prematurely when it encounters infeasible solutions to EVRP-TW-SPD by moving in the descent direction of the VRPSPDTW objective value, potentially missing many feasible solutions in the EVRP-TW-SPD neighborhood. 
% Therefore, we also need CLS that strictly performs move evaluation in the EVRP-TW-SPD neighborhood~(lines~\ref{algline: CDNS REPEAT2}--\ref{algline: CDNS UNTIL2}). 

\subsubsection{Discussion}
By exploring two distinct solution spaces, CNDS typically finds higher-quality solutions than using either solution space alone, as demonstrated in our experiments (see Section~\ref{subsec: Effectiveness of Each Component in HMA}).
The idea of CDNS can be readily generalized to other EVRP variants by simultaneously exploring both electric and non-electric neighborhoods.
This dual-domain approach is particularly valuable since non-electric variants are typically more extensively studied in the literature, providing a rich pool of well-established local search procedures that can be adapted for solving various EVRPs efficiently.

% This local search design concept can be easily generalized to other EVRP variants. 
% In other words, when encountering a new EVRP variant, it is beneficial to first review the non-electric
% counterpart and its state-of-the-art algorithms. 
% Then, as Proposition~\ref{proposition: VRP-SPD-SPD & EVRP-TW-SPD} illustrates, understanding the relationship between the solutions of these two problems contributes to the slight modification and easy reuse of non-electric
% counterpart VRP variant methods to solve the new EVRP variant efficiently.

\subsection{Overall Framework of HMA}
\label{sec:HMA}
 
 \begin{figure}[tbp]
	\centering
	\scalebox{1.0}{\includegraphics[width=\linewidth]{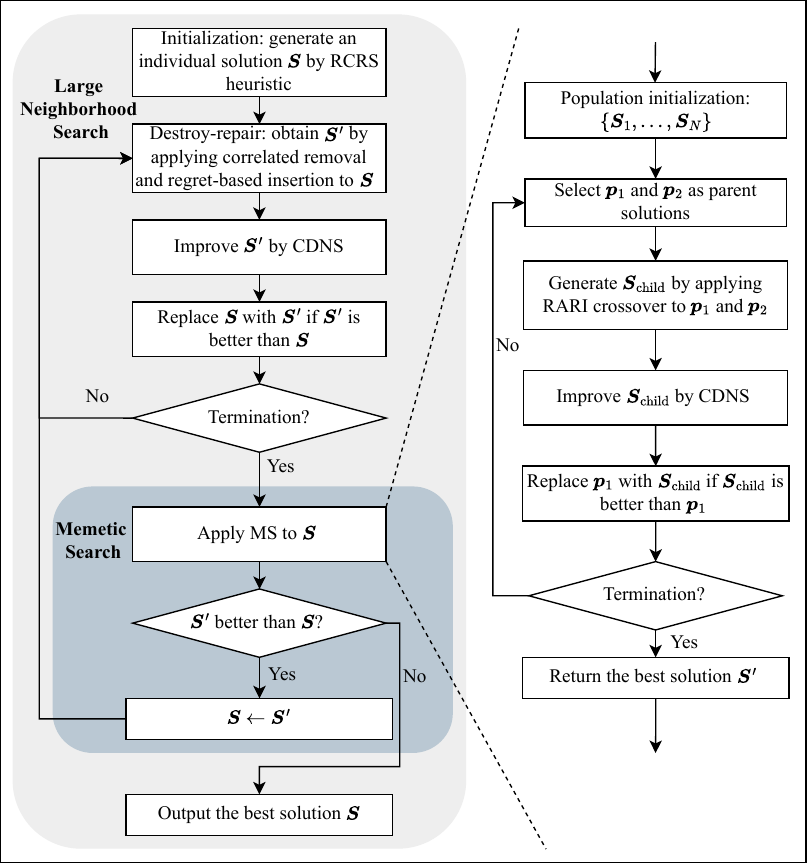}}
	\caption{\textcolor{black}{The flow chart of HMA.}}
	\label{fig:flow_chart}
\end{figure}

We now introduce the framework of HMA, which integrates an individual-based large neighborhood search~(LNS) and a population-based memetic search~(MS).
The flow chart of HMA is illustrated in Figure \ref{fig:flow_chart}.
Overall, HMA employs LNS to efficiently obtain a high-quality solution, and then uses MS to further improve it.

As presented in Algorithm~\ref{alg: Hybrid Search Framework}, HMA first enters the individual-based LNS.
It begins with the initialization of an individual solution~$\bm{S}$, which is constructed using the RCRS heuristic~\cite{dethloff2001vehicle}.
% and best station insertion~\cite{keskin2016partial}.
Specifically, RCRS is an extension of the cheapest-insertion heuristic and constructs routes by iteratively inserting an unassigned customer at the position with the minimal value according to the RCRS criterion until no feasible insertions are possible due to capacity or time window constraints (please refer to~\cite{dethloff2001vehicle} for details of RCRS).
The RCRS criterion combines the total travel distance, residual capacity, and radial surcharge with weighting parameters $\lambda, \gamma \in [0, 1]$.
% to construct a high-quality solution for handling simultaneous pickup-delivery.
Here, we simply set $\lambda=\gamma=0.5$.
In this initialization, electricity-feasible customers are first considered for insertion, followed by those that are only electricity-infeasible but can be handled later through best station insertion.
When no feasible insertions remain, a new route is initiated. 
This process continues until all customers are assigned~(line~\ref{algline: HS initialization}).

% Besides the conventional cheapest-insertion heuristic's total travel distance~(TD), the \textit{RCRS} criterion combines the residual capacity~(RC) and the radial surcharge~(RS), with weighting parameters $\lambda, \gamma \in [0, 1]$, so that it can construct a high-quality initial solution for handling the SPD problem. 
% In this initialization, we simply set $\lambda=\gamma=0.5$~(line~\ref{algline: HS initialization}). 

\begin{algorithm}[tbp]
    \small
    \caption{HMA}
    \label{alg: Hybrid Search Framework}
    \LinesNumbered
    \KwIn{an EVRP-TW-SPD instance: \( \bm{G} = (\bm{V}', \bm{E}) \)} \label{algline: HS requirement}
    \KwOut{the best-found solution: $\bm{S}$} \label{algline: HS ensure}

    \tcc{Large Neighborhood Search~(LNS)}
    $\{\bm{S}\} \gets \text{Initialization}()$\; \label{algline: HS initialization}
    \While{true}{  \label{algline: HS while}
        $\bm{S}' \gets \bm{S}$\;
        remove $\rho$ customers from $\bm{S}'$ and reinsert them into $\bm{S}'$ via regret-based insertion\; \label{algline: HS destroy and repair}
        $\bm{S}' \gets \text{CDNS}(\bm{S}')$\;  \label{algline: HS CDNS1}
        \lIf{$\bm{S}'$ is better than $\bm{S}$}{$\bm{S} \gets \bm{S}'$} \label{algline: HS Select Best}
        
        \If{$\bm{S}$ is not improved in the last $G_1$ iterations}{ \label{algline: HS if check}
            \tcc{Memetic Search~(MS)}
            $\bm{S}' \gets \bm{S}$\; \label{algline: enter}
            $\{\bm{S}_{1},\ldots,\bm{S}_{N}\} \gets \text{Population\_Initialization}(\bm{S}')$\; \label{algline: HS population introduction}
            \Repeat{$\bm{S}'$ is not improved in the last $G_2$ generations}{ \label{algline: HS repeat inner}
                $\pi(\cdot) \gets \text{a random permutation of } 1, \ldots, N$\; \label{algline: HS random permutation}
                \For{$i \gets 1$ \KwTo $N$}{ \label{algline: HS for N}
                    $\bm{p}_1 \gets \bm{S}_{\pi(i)},\ \bm{p}_2 \gets \bm{S}_{\pi(i+1)}$\; \label{algline: HS parents}
                    $\bm{S}_{\text{child}} \gets \text{Crossover}(\bm{p}_1, \bm{p}_2)$\; \label{algline: HS crossover}
                    $\bm{S}_{\text{child}} \gets$ CDNS$(\bm{S}_{\text{child}})$\; \label{algline: HS Cross-Domain Neighborhood Search}
                    $\bm{S}_{\pi(i)} \gets \text{Select\_Best}(\bm{S}_{\text{child}}, \bm{p}_1)$\; \label{algline: HS select best child}
                } \label{algline: HS end for N}
                $\bm{S}' \gets \text{Select\_Best}(\bm{S}' ,\bm{S}_{1},\ldots,\bm{S}_{N})$\; \label{algline: HS select best population}
            } \label{algline: HS until improvement}
            \lIf{$\bm{S}'$ is better than $\bm{S}$}{$\bm{S} \gets \bm{S}'$}  \label{algline: HS better}
		  \lElse{break}  \label{algline: HS break}
        } \label{algline: HS end if}
    } \label{algline: HS endwhile}
    \Return{$\bm{S}$} \label{algline: HS return}
\end{algorithm}

After initialization, HMA applies the destroy-repair operator as described in \cite{liu2021memetic} (a large-step perturbation) on the individual solution~$\bm{S}$ (line~\ref{algline: HS destroy and repair}).
This destroy-repair operator involves removing $\rho$ correlated customers from the solution, where the random number $\rho \in [\omega_1|\bm{V}|, \omega_2|\bm{V}|]$, and $\omega_1$, $\omega_2$ are parameters with $0 < \omega_1 \leq \omega_2 < 1$.
The operator then reinserts these customers using a regret-based heuristic that evaluates the expected cost of inserting a node in a future iteration rather than in the current one to assess the quality of potential insertions.
% This destroy-repair involves removing $\rho$ correlated customers from $\bm{S}$, where random number $\rho \in [\omega_1|\bm{V}|, \omega_2|\bm{V}|]$, 
% $\omega_1$, $\omega_2$ are parameters with $0 < \omega_1 \leq \omega_2 < 1$ (line~\ref{algline: HS destroy}), and then reinserting them using a regret-based heuristic that evaluates the expected cost of inserting a node in a future iteration rather than in the current one to assess the quality of potential insertions (line~\ref{algline: HS repair}). 
The process is followed by CDNS, ensuring that a new local optimum is reached to potentially improve $\bm{S}$~(lines~\ref{algline: HS CDNS1}--\ref{algline: HS Select Best}).
If $\bm{S}$ does not show improvement over a specified number of consecutive iterations, denoted as $G_1$, HMA initiates the population-based MS~(lines~\ref{algline: enter}--\ref{algline: HS until improvement}).

MS starts with a population of $N$ individuals $\bm{S}_1,...,\bm{S}_N$, generated based on the current elite solution $\bm{S}$~(line~\ref{algline: HS population introduction}).
Specifically, $\bm{S}_{1}$ is set as a copy of $\bm{S}$ because the elite solution should be preserved. 
As for the remaining individuals, the individual $\bm{S}_{i}$ with odd index $i$ comes from the destroy-repair operator performed on $\bm{S}$ with $\omega_1=\omega_2=i/N$;
for the individual $\bm{S}_{i}$ with even index $i$, it is generated by RCRS with $\lambda = \frac{1}{\sqrt{N}-1}\cdot(p-1), \gamma = \frac{1}{\sqrt{N}-1}\cdot(q-1)$, where $p = \left\lceil i/\sqrt{N} \right\rceil$ and $q=i-\sqrt{N} \cdot(p-1)$.
The above procedure would result in a diverse population of individuals while preserving the high-quality information of the elite solution obtained from the previous LNS.

% More details about the evolutionary process of MS(lines~\ref{algline: HS repeat inner}--\ref{algline: HS until improvement}) can be seen in~\cite{liu2021memetic}; we do not describe it further in this paper.

After initialization, MS enters an evolutionary process (lines~\ref{algline: HS repeat inner}--\ref{algline: HS until improvement}).
In each generation (lines~\ref{algline: HS for N}--\ref{algline: HS end for N}), each individual in the population is selected once as parent $\bm{p}_1$ and once as parent $\bm{p}_2$, in a random order (lines~\ref{algline: HS random permutation} and \ref{algline: HS parents}).
For each pair of parent solutions, MS generates an offspring using the route-assembly-regret-insertion (RARI) crossover~\cite{liu2021memetic} (line~\ref{algline: HS crossover}), and then improves it through CDNS (line~\ref{algline: HS Cross-Domain Neighborhood Search}).
After that, the better one among the oﬀspring and $\bm{p}_1$ replaces the individual selected as $\bm{p}_1$ (line~\ref{algline: HS select best child}).
After MS, a new candidate solution $\bm{S}'$ is obtained with the aim of improving $\bm{S}$ (lines~\ref{algline: HS better}--\ref{algline: HS break}).
If the quality of $\bm{S}$ is still not improved by MS, then HMA would output $\bm{S}$ and terminate (line~\ref{algline: HS break}); otherwise, HMA proceeds to the next iteration of LNS~(line~\ref{algline: HS better}).

When presented with a large-scale problem instance (with more than 200 customers) to solve, we make two adjustments to HMA to improve its efficiency.
First, only ALS is used in CDNS.
Second, problem decomposition is employed in MS.
Specifically,  the decomposition breaks down the main problem~$\bm{G} = (\bm{V}', \bm{E})$ into $k$ smaller subproblems~$\bm{G}_1, \ldots, \bm{G}_k$ when entering MS~(line~\ref{algline: enter}).
A subproblem $\bm{G}_i = (\bm{V}_i', \bm{E}_i) $ with  $\bm{V}_i' = \{0\} \cup \bm{V}_i \cup \bm{F}$, $\bm{E}_i = \{ (p, q) | p, q \in \bm{V}_i', p \neq q \}$, where $\bm{V} = \bigcup_{i=1}^k \bm{V}_i  \text{ and }  \bm{V}_i \cap \bm{V}_j = \emptyset \ \forall i \neq j,\ 1 \leq i,j \leq k$.
Here, we use the barycenter clustering decomposition~(BCD)~\cite{santini2023decomposition} to determine the partition of $\bm{V}_i$.
The barycenter of a route $\bm{R} = (n_{1}, n_{2}, \ldots, n_{L})$ is defined as $(\overline{x}, \overline{y}) = \frac{1}{L-1} \sum_{i=1}^{L-1} (x_i, y_i),$
where $(x_i, y_i)$ is the Cartesian plane coordinate of node $n_i$.
For each subproblem $\bm{G}_i$, the same MS is applied~(lines~\ref{algline: HS population introduction}--\ref{algline: HS until improvement}).
After all MS procedures are finished, the solutions from all subproblems are assembled to construct a complete solution $\bm{S}'$ for the original problem.

\begin{table}[tbp]
\centering
\caption{Setting of algorithm parameters.}
\label{tab: parameter_setting}
\renewcommand{\arraystretch}{1.2}
\resizebox{\columnwidth}{!}{%
\begin{tabular}{ccccccccc}
\toprule
Parameter & $G_1$ & $G_2$ & $N$ & $\alpha$ & $B$ & $sr$  & $w_1, w_2$ \\ 
\midrule
$akb_{small}$  & 20 & 20 & 9 & 3 & 5 & $1.0$ & $0.2, 0.4$ \\ 
$akb_{medium}$   & 20 & 20 & 4 & 3 & 5 & $0.5$ & $0.1, 0.2$ \\ 
$jd$ & 20 & 20 & 4 & 3 & 5 & $0.1$ & $0.05, 0.05$ \\
\bottomrule
\end{tabular}%
}
\end{table}

\section{Experiments}
\label{sec: Computational Study}
To evaluate the effectiveness of HMA, we compared it with the recent state-of-the-art algorithms for solving EVRP-TW-SPD, Adapt-CMSA-STD~\cite{akbay2022application} and Adapt-CMSA-SETCOV~\cite{akbay2023cmsa}, using the existing benchmark set.
Then, we introduced a new benchmark set derived from a real-world application that includes large-scale instances.
The evaluation results of Adapt-CMSA-STD, Adapt-CMSA-SETCOV, and HMA on the new benchmark were also reported.
Moreover, we used a statistical test to support our claims and measure algorithm robustness.
Finally, we conducted a comprehensive ablation study to assess the effectiveness of PSSI, CDNS, and the hybrid search framework. 
The source code of HMA and the new benchmark set are open-sourced at \href{https://anonymous.4open.science/r/EVRP-TW-SPD-HMA-code-dataset}{https://anonymous.4open.science/r/EVRP-TW-SPD-HMA-code-dataset.}

\begin{table*}[tbp]
\caption{Comparative results on small-scale instances in the \( akb \) set.}
\label{tab: small-scale}
\resizebox{\textwidth}{!}{%
\begin{tabular}{cccclcccclcccclccccc}
\toprule
 & \multicolumn{3}{c}{CPLEX} &  & \multicolumn{4}{c}{Adapt-CMSA-STD} &  & \multicolumn{4}{c}{Adapt-CMSA-SETCOV} &  & \multicolumn{5}{c}{HMA}  \\ \cline{2-4} \cline{6-9} \cline{11-14} \cline{16-20}
\multirow{-2}{*}{Instance name} & m & best & time &  & m & best & avg & time &  & m & best & avg & time &  & m & best & avg & time & gap \\ 
\midrule
c101C5 & 2 & \cellcolor[HTML]{D9D9D9}2257.75 & 0.43 &  & 2 & \cellcolor[HTML]{D9D9D9}2257.75 & \cellcolor[HTML]{D9D9D9}2257.75 & 0.02 &  & 2 & \cellcolor[HTML]{D9D9D9}2257.75 & \cellcolor[HTML]{D9D9D9}2257.75 & 0.01 &  & 2 & \cellcolor[HTML]{D9D9D9}2257.75 & \cellcolor[HTML]{D9D9D9}2257.75 & 0.38 & 0.00\% \\
c103C5 & 1 & \cellcolor[HTML]{D9D9D9}1175.37 & 0.41 &  & 1 & \cellcolor[HTML]{D9D9D9}1175.37 & \cellcolor[HTML]{D9D9D9}1175.37 & 0.54 &  & 1 & \cellcolor[HTML]{D9D9D9}1175.37 & \cellcolor[HTML]{D9D9D9}1175.37 & 0.68 &  & 1 & \cellcolor[HTML]{D9D9D9}1175.37 & \cellcolor[HTML]{D9D9D9}1175.37 & 0.21 & 0.00\% \\
c206C5 & 1 & \cellcolor[HTML]{D9D9D9}1242.56 & 0.58 &  & 1 & \cellcolor[HTML]{D9D9D9}1242.56 & \cellcolor[HTML]{D9D9D9}1242.56 & 0.03 &  & 1 & \cellcolor[HTML]{D9D9D9}1242.56 & \cellcolor[HTML]{D9D9D9}1242.56 & 0.00 &  & 1 & \cellcolor[HTML]{D9D9D9}1242.56 & \cellcolor[HTML]{D9D9D9}1242.56 & 0.19 & 0.00\% \\
c208C5 & 1 & \cellcolor[HTML]{D9D9D9}1158.48 & 0.08 &  & 1 & \cellcolor[HTML]{D9D9D9}1158.48 & \cellcolor[HTML]{D9D9D9}1158.48 & 0.01 &  & 1 & \cellcolor[HTML]{D9D9D9}1158.48 & \cellcolor[HTML]{D9D9D9}1158.48 & 0.00 &  & 1 & \cellcolor[HTML]{D9D9D9}1158.48 & \cellcolor[HTML]{D9D9D9}1158.48 & 0.24 & 0.00\% \\
r104C5 & 2 & \cellcolor[HTML]{D9D9D9}2136.69 & 0.02 &  & 2 & \cellcolor[HTML]{D9D9D9}2136.69 & \cellcolor[HTML]{D9D9D9}2136.69 & 0.07 &  & 2 & \cellcolor[HTML]{D9D9D9}2136.69 & \cellcolor[HTML]{D9D9D9}2136.69 & 0.01 &  & 2 & \cellcolor[HTML]{D9D9D9}2136.69 & \cellcolor[HTML]{D9D9D9}2136.69 & 0.93 & 0.00\% \\
r105C5 & 2 & \cellcolor[HTML]{D9D9D9}2156.08 & 0.03 &  & 2 & \cellcolor[HTML]{D9D9D9}2156.08 & \cellcolor[HTML]{D9D9D9}2156.08 & 0.01 &  & 2 & \cellcolor[HTML]{D9D9D9}2156.08 & \cellcolor[HTML]{D9D9D9}2156.08 & 0.00 &  & 2 & \cellcolor[HTML]{D9D9D9}2156.08 & \cellcolor[HTML]{D9D9D9}2156.08 & 1.28 & 0.00\% \\
r202C5 & 1 & \cellcolor[HTML]{D9D9D9}1128.78 & 0.06 &  & 1 & \cellcolor[HTML]{D9D9D9}1128.78 & \cellcolor[HTML]{D9D9D9}1128.78 & 9.00 &  & 1 & \cellcolor[HTML]{D9D9D9}1128.78 & \cellcolor[HTML]{D9D9D9}1128.78 & 0.00 &  & 1 & \cellcolor[HTML]{D9D9D9}1128.78 & \cellcolor[HTML]{D9D9D9}1128.78 & 0.24 & 0.00\% \\
r203C5 & 1 & \cellcolor[HTML]{D9D9D9}1179.06 & 0.03 &  & 1 & \cellcolor[HTML]{D9D9D9}1179.06 & \cellcolor[HTML]{D9D9D9}1179.06 & 0.22 &  & 1 & \cellcolor[HTML]{D9D9D9}1179.06 & \cellcolor[HTML]{D9D9D9}1179.06 & 0.05 &  & 1 & \cellcolor[HTML]{D9D9D9}1179.06 & \cellcolor[HTML]{D9D9D9}1179.06 & 0.17 & 0.00\% \\
rc105C5 & 2 & \cellcolor[HTML]{D9D9D9}2233.77 & 2.18 &  & 2 & \cellcolor[HTML]{D9D9D9}2233.77 & \cellcolor[HTML]{D9D9D9}2233.77 & 0.09 &  & 2 & \cellcolor[HTML]{D9D9D9}2233.77 & \cellcolor[HTML]{D9D9D9}2233.77 & 0.04 &  & 2 & \cellcolor[HTML]{D9D9D9}2233.77 & \cellcolor[HTML]{D9D9D9}2233.77 & 0.86 & 0.00\% \\
rc108C5 & 2 & \cellcolor[HTML]{D9D9D9}2253.93 & 0.19 &  & 2 & \cellcolor[HTML]{D9D9D9}2253.93 & \cellcolor[HTML]{D9D9D9}2253.93 & 0.01 &  & 2 & \cellcolor[HTML]{D9D9D9}2253.93 & \cellcolor[HTML]{D9D9D9}2253.93 & 0.00 &  & 2 & \cellcolor[HTML]{D9D9D9}2253.93 & \cellcolor[HTML]{D9D9D9}2253.93 & 0.58 & 0.00\% \\
rc204C5 & 1 & \cellcolor[HTML]{D9D9D9}1176.39 & 0.25 &  & 1 & \cellcolor[HTML]{D9D9D9}1176.39 & \cellcolor[HTML]{D9D9D9}1176.39 & 0.07 &  & 1 & \cellcolor[HTML]{D9D9D9}1176.39 & \cellcolor[HTML]{D9D9D9}1176.39 & 0.01 &  & 1 & \cellcolor[HTML]{D9D9D9}1176.39 & \cellcolor[HTML]{D9D9D9}1176.39 & 0.21 & 0.00\% \\
rc208C5 & 1 & \cellcolor[HTML]{D9D9D9}1167.98 & 0.12 &  & 1 & \cellcolor[HTML]{D9D9D9}1167.98 & \cellcolor[HTML]{D9D9D9}1167.98 & 0.52 &  & 1 & \cellcolor[HTML]{D9D9D9}1167.98 & \cellcolor[HTML]{D9D9D9}1167.98 & 0.03 &  & 1 & \cellcolor[HTML]{D9D9D9}1167.98 & \cellcolor[HTML]{D9D9D9}1167.98 & 0.19 & 0.00\% \\
c101C10 & 3 & \cellcolor[HTML]{D9D9D9}3388.25 & 76.74 &  & 3 & \cellcolor[HTML]{D9D9D9}3388.25 & \cellcolor[HTML]{D9D9D9}3388.25 & 0.28 &  & 3 & \cellcolor[HTML]{D9D9D9}3388.25 & 3388.55 & 0.33 &  & 3 & \cellcolor[HTML]{D9D9D9}3388.25 & \cellcolor[HTML]{D9D9D9}3388.25 & 1.70 & 0.00\% \\
c104C10 & 2 & \cellcolor[HTML]{D9D9D9}2273.93 & 2.16 &  & 2 & \cellcolor[HTML]{D9D9D9}2273.93 & \cellcolor[HTML]{D9D9D9}2273.93 & 0.48 &  & 2 & \cellcolor[HTML]{D9D9D9}2273.93 & \cellcolor[HTML]{D9D9D9}2273.93 & 29.02 &  & 2 & \cellcolor[HTML]{D9D9D9}2273.93 & \cellcolor[HTML]{D9D9D9}2273.93 & 1.10 & 0.00\% \\
c202C10 & 1 & \cellcolor[HTML]{D9D9D9}1304.06 & 6.34 &  & 1 & \cellcolor[HTML]{D9D9D9}1304.06 & \cellcolor[HTML]{D9D9D9}1304.06 & 0.45 &  & 1 & \cellcolor[HTML]{D9D9D9}1304.06 & \cellcolor[HTML]{D9D9D9}1304.06 & 0.10 &  & 1 & \cellcolor[HTML]{D9D9D9}1304.06 & \cellcolor[HTML]{D9D9D9}1304.06 & 1.40 & 0.00\% \\
c205C10 & 2 & \cellcolor[HTML]{D9D9D9}2228.28 & 0.08 &  & 2 & \cellcolor[HTML]{D9D9D9}2228.28 & \cellcolor[HTML]{D9D9D9}2228.28 & 10.82 &  & 2 & \cellcolor[HTML]{D9D9D9}2228.28 & \cellcolor[HTML]{D9D9D9}2228.28 & 0.03 &  & 2 & \cellcolor[HTML]{D9D9D9}2228.28 & \cellcolor[HTML]{D9D9D9}2228.28 & 0.45 & 0.00\% \\
r102C10 & 3 & \cellcolor[HTML]{D9D9D9}3249.19 & 0.56 &  & 3 & \cellcolor[HTML]{D9D9D9}3249.19 & \cellcolor[HTML]{D9D9D9}3249.19 & 18.22 &  & 3 & \cellcolor[HTML]{D9D9D9}3249.19 & \cellcolor[HTML]{D9D9D9}3249.19 & 0.01 &  & 3 & \cellcolor[HTML]{D9D9D9}3249.19 & \cellcolor[HTML]{D9D9D9}3249.19 & 1.03 & 0.00\% \\
r103C10 & 2 & \cellcolor[HTML]{D9D9D9}2206.12 & 17.49 &  & 2 & \cellcolor[HTML]{D9D9D9}2206.12 & \cellcolor[HTML]{D9D9D9}2206.12 & 5.27 &  & 2 & \cellcolor[HTML]{D9D9D9}2206.12 & \cellcolor[HTML]{D9D9D9}2206.12 & 22.01 &  & 2 & \cellcolor[HTML]{D9D9D9}2206.12 & \cellcolor[HTML]{D9D9D9}2206.12 & 3.38 & 0.00\% \\
r201C10 & 1 & \cellcolor[HTML]{D9D9D9}1241.51 & 61.20 &  & 1 & \cellcolor[HTML]{D9D9D9}1241.51 & \cellcolor[HTML]{D9D9D9}1241.51 & 22.71 &  & 1 & \cellcolor[HTML]{D9D9D9}1241.51 & \cellcolor[HTML]{D9D9D9}1241.51 & 15.97 &  & 1 & \cellcolor[HTML]{D9D9D9}1241.51 & \cellcolor[HTML]{D9D9D9}1241.51 & 3.29 & 0.00\% \\
r203C10 & 1 & \cellcolor[HTML]{D9D9D9}1218.21 & 1.57 &  & 1 & \cellcolor[HTML]{D9D9D9}1218.21 & \cellcolor[HTML]{D9D9D9}1218.21 & 14.06 &  & 1 & \cellcolor[HTML]{D9D9D9}1218.21 & \cellcolor[HTML]{D9D9D9}1218.21 & 27.17 &  & 1 & \cellcolor[HTML]{D9D9D9}1218.21 & \cellcolor[HTML]{D9D9D9}1218.21 & 0.97 & 0.00\% \\
rc102C10 & 4 & \cellcolor[HTML]{D9D9D9}4423.51 & 0.15 &  & 4 & \cellcolor[HTML]{D9D9D9}4423.51 & \cellcolor[HTML]{D9D9D9}4423.51 & 0.24 &  & 4 & \cellcolor[HTML]{D9D9D9}4423.51 & \cellcolor[HTML]{D9D9D9}4423.51 & 0.01 &  & 4 & \cellcolor[HTML]{D9D9D9}4423.51 & \cellcolor[HTML]{D9D9D9}4423.51 & 3.27 & 0.00\% \\
rc108C10 & 3 & \cellcolor[HTML]{D9D9D9}3345.93 & 3.72 &  & 3 & \cellcolor[HTML]{D9D9D9}3345.93 & \cellcolor[HTML]{D9D9D9}3345.93 & 14.61 &  & 3 & \cellcolor[HTML]{D9D9D9}3345.93 & \cellcolor[HTML]{D9D9D9}3345.93 & 0.01 &  & 3 & \cellcolor[HTML]{D9D9D9}3345.93 & \cellcolor[HTML]{D9D9D9}3345.93 & 1.29 & 0.00\% \\
rc201C10 & 1 & \cellcolor[HTML]{D9D9D9}1412.86 & 22756.49 &  & 1 & \cellcolor[HTML]{D9D9D9}1412.86 & \cellcolor[HTML]{D9D9D9}1412.86 & 1.97 &  & 1 & \cellcolor[HTML]{D9D9D9}1412.86 & \cellcolor[HTML]{D9D9D9}1412.86 & 0.93 &  & 1 & \cellcolor[HTML]{D9D9D9}1412.86 & \cellcolor[HTML]{D9D9D9}1412.86 & 6.03 & 0.00\% \\
rc205C10 & 2 & \cellcolor[HTML]{D9D9D9}2325.98 & 0.13 &  & 2 & \cellcolor[HTML]{D9D9D9}2325.98 & \cellcolor[HTML]{D9D9D9}2325.98 & 0.13 &  & 2 & \cellcolor[HTML]{D9D9D9}2325.98 & \cellcolor[HTML]{D9D9D9}2325.98 & 0.45 &  & 2 & \cellcolor[HTML]{D9D9D9}2325.98 & \cellcolor[HTML]{D9D9D9}2325.98 & 0.74 & 0.00\% \\
c103C15 & 3 & \cellcolor[HTML]{D9D9D9}3348.46 & 5046.46 &  & 3 & \cellcolor[HTML]{D9D9D9}3348.46 & 3348.47 & 48.19 &  & 3 & \cellcolor[HTML]{D9D9D9}3348.46 & \cellcolor[HTML]{D9D9D9}3348.46 & 3.76 &  & 3 & \cellcolor[HTML]{D9D9D9}3348.46 & \cellcolor[HTML]{D9D9D9}3348.46 & 3.30 & 0.00\% \\
c106C15 & 3 & \cellcolor[HTML]{D9D9D9}3275.13 & 0.90 &  & 3 & \cellcolor[HTML]{D9D9D9}3275.13 & \cellcolor[HTML]{D9D9D9}3275.13 & 4.56 &  & 3 & \cellcolor[HTML]{D9D9D9}3275.13 & \cellcolor[HTML]{D9D9D9}3275.13 & 32.34 &  & 3 & \cellcolor[HTML]{D9D9D9}3275.13 & \cellcolor[HTML]{D9D9D9}3275.13 & 2.42 & 0.00\% \\
c202C15 & 2 & \cellcolor[HTML]{D9D9D9}2383.62 & 43.74 &  & 2 & \cellcolor[HTML]{D9D9D9}2383.62 & \cellcolor[HTML]{D9D9D9}2383.62 & 12.73 &  & 2 & \cellcolor[HTML]{D9D9D9}2383.62 & 2393.39 & 5.27 &  & 2 & \cellcolor[HTML]{D9D9D9}2383.62 & \cellcolor[HTML]{D9D9D9}2383.62 & 1.67 & 0.00\% \\
c208C15 & 2 & \cellcolor[HTML]{D9D9D9}2300.55 & 3.25 &  & 2 & \cellcolor[HTML]{D9D9D9}2300.55 & \cellcolor[HTML]{D9D9D9}2300.55 & 1.09 &  & 2 & \cellcolor[HTML]{D9D9D9}2300.55 & \cellcolor[HTML]{D9D9D9}2300.55 & 8.59 &  & 2 & \cellcolor[HTML]{D9D9D9}2300.55 & \cellcolor[HTML]{D9D9D9}2300.55 & 1.36 & 0.00\% \\
r102C15 & 5 & \cellcolor[HTML]{D9D9D9}5412.78 & 5046.62 &  & 5 & \cellcolor[HTML]{D9D9D9}5412.78 & \cellcolor[HTML]{D9D9D9}5412.78 & 1.82 &  & 5 & \cellcolor[HTML]{D9D9D9}5412.78 & \cellcolor[HTML]{D9D9D9}5412.78 & 0.09 &  & 5 & \cellcolor[HTML]{D9D9D9}5412.78 & \cellcolor[HTML]{D9D9D9}5412.78 & 4.27 & 0.00\% \\
r105C15 & 4 & \cellcolor[HTML]{D9D9D9}4336.15 & 5.34 &  & 4 & \cellcolor[HTML]{D9D9D9}4336.15 & \cellcolor[HTML]{D9D9D9}4336.15 & 1.03 &  & 4 & \cellcolor[HTML]{D9D9D9}4336.15 & \cellcolor[HTML]{D9D9D9}4336.15 & 0.86 &  & 4 & \cellcolor[HTML]{D9D9D9}4336.15 & \cellcolor[HTML]{D9D9D9}4336.15 & 2.03 & 0.00\% \\
r202C15 & 2 & 2361.51 & 5045.36 &  & 2 & 2358.00 & 2364.90 & 22.71 &  & 1 & \cellcolor[HTML]{D9D9D9}1507.32 & \cellcolor[HTML]{D9D9D9}1677.46 & 45.16 &  & 2 & 2358.00 & 2358.00 & 4.00 & 56.44\% \\
r209C15 & 1 & \cellcolor[HTML]{D9D9D9}1313.24 & 3088.27 &  & 1 & \cellcolor[HTML]{D9D9D9}1313.24 & \cellcolor[HTML]{D9D9D9}1313.24 & 12.23 &  & 1 & \cellcolor[HTML]{D9D9D9}1313.24 & \cellcolor[HTML]{D9D9D9}1313.24 & 10.98 &  & 1 & \cellcolor[HTML]{D9D9D9}1313.24 & \cellcolor[HTML]{D9D9D9}1313.24 & 2.83 & 0.00\% \\
rc103C15 & 4 & \cellcolor[HTML]{D9D9D9}4397.67 & 245.61 &  & 4 & \cellcolor[HTML]{D9D9D9}4397.67 & \cellcolor[HTML]{D9D9D9}4397.67 & 0.24 &  & 4 & \cellcolor[HTML]{D9D9D9}4397.67 & \cellcolor[HTML]{D9D9D9}4397.67 & 0.21 &  & 4 & \cellcolor[HTML]{D9D9D9}4397.67 & \cellcolor[HTML]{D9D9D9}4397.67 & 2.73 & 0.00\% \\
rc108C15 & 3 & \cellcolor[HTML]{D9D9D9}3370.25 & 822.47 &  & 3 & \cellcolor[HTML]{D9D9D9}3370.25 & \cellcolor[HTML]{D9D9D9}3370.25 & 41.34 &  & 3 & \cellcolor[HTML]{D9D9D9}3370.25 & \cellcolor[HTML]{D9D9D9}3370.25 & 0.24 &  & 3 & \cellcolor[HTML]{D9D9D9}3370.25 & \cellcolor[HTML]{D9D9D9}3370.25 & 1.87 & 0.00\% \\
rc202C15 & 2 & \cellcolor[HTML]{D9D9D9}2394.39 & 603.76 &  & 2 & \cellcolor[HTML]{D9D9D9}2394.39 & \cellcolor[HTML]{D9D9D9}2394.39 & 0.48 &  & 2 & \cellcolor[HTML]{D9D9D9}2394.39 & \cellcolor[HTML]{D9D9D9}2394.39 & 19.65 &  & 2 & \cellcolor[HTML]{D9D9D9}2394.39 & \cellcolor[HTML]{D9D9D9}2394.39 & 2.72 & 0.00\% \\
rc204C15 & 1 & 1403.38 & 5046.60 &  & 1 & \cellcolor[HTML]{D9D9D9}1382.22 & 1385.72 & 47.83 &  & 1 & \cellcolor[HTML]{D9D9D9}1382.22 & \cellcolor[HTML]{D9D9D9}1382.55 & 50.49 &  & 1 & \cellcolor[HTML]{D9D9D9}1382.22 & 1384.83 & 19.75 & 0.00\% \\ 
\midrule
avg & 2.06 & 2338.38 & 1331.37 &  & 2.06 & 2337.70 & 2337.99 & 8.17 &  & 2.03 & 2314.07 & 2319.08 & 7.62 &  & 2.06 & 2337.70 & 2337.77 & 2.20 & 1.57\% \\
\bottomrule
\end{tabular}%
}
\end{table*}

\subsection{Experimental Setup}
\subsubsection{Benchmark}
\label{benchmark}
As noted by Akbay~\textit{et~al.}\cite{akbay2022application}, the existing benchmark used is based on E-VRPTW instances from Schneider~\textit{et~al.}\cite{schneider2014electric}, with customer demands separated into delivery and pickup demands using the method in \cite{salhi1999cluster}. 
We refer to this benchmark as the~\textit{akb} (authors’ initials) set. 
This~\textit{akb} set has 92 instances, including 36 small-scale instances~(twelve 5-customer instances, twelve 10-customer instances, and twelve 15-customer instances, each with 2 to 8 stations) and 56 medium-scale instances~(100-customer instances with 21 stations ). 
These instances are categorized into three groups based on the spatial distribution of customer locations:
clustered (prefixed with~\textit{c}),
randomly distributed (prefixed with~\textit{r}),
and a mixture of random and clustered distributions (prefixed with~\textit{rc}). 
Each group is further divided into two sub-classes:
type 1 features narrow time windows and small vehicle and battery capacities,
while type 2 has wide time windows and large vehicle and battery capacities.
% Each group is further divided into two sub-classes: type~1 features narrow time windows, smaller vehicle and battery capacities, while type~2 has larger time windows, greater vehicle and battery capacities. 
All these instances are defined in a two-dimensional Euclidean space.
That is, each node $i$ has the Cartesian coordinate $(x_i, y_i)$. 
The distance between two nodes $i$ and $j$ is given by the Euclidean distance $\sqrt{(x_i - x_j)^2 + (y_i - y_j)^2}$, which is a special case of the problem formulation provided in this paper (see Section~\ref{sec: Notations and Problem Definition}).
% where the distances between nodes are explicitly specified regardless of whether they are defined in the Euclidean plane. 
All distances and time between two nodes $i$ and $j$ satisfy the triangle inequality.
Following \cite{akbay2022application,akbay2023cmsa}, in the experiments conducted on the \textit{akb} set, we set $\mu_1=1000$ and $\mu_2=1$.

As previously mentioned, the~\textit{akb} set includes only synthetic instances of small and medium scales. 
To address the gap in benchmark instance scales, we introduced a new benchmark set derived from JD Logistics' distribution system. 
% The raw data was collected from this system, which contains requests occurring during a period of time in Beijing.
% In addition to delivering goods purchased by customers, it also involves collecting goods from customers.
The raw data collected from this system includes requests over a period in Beijing, involving both deliveries of purchased goods and collections picked up from customers.
We sampled from this data to generate instances with 200, 400, 600, 800, and 1000 customers, each with 100 stations. 
For each problem scale, we generated 4 instances, resulting in a set of 20 large-scale instances, referred to as the~\textit{jd} set.
Since the~\textit{jd} dataset is derived from real-world data, each instance provides the longitude and latitude $(lng_i, lat_i)$ of each node $i$, in degrees. 
% However, because the area covered is not large, we can use Mercator Projection to convert $(lng_i, lat_i)$ into Cartesian plane coordinate $(x_i, y_i)$, i.e., $x_i = R \cdot \frac{\pi \cdot lng}_i{180}$, $y_i = R \cdot \ln \left( \tan \left( \frac{\pi}{4} + \frac{\pi \cdot lat}_i{360} \right) \right)$, where $R$ is the semi-major axis of the WGS84 ellipsoid model. 
Additionally, the distances and time between two nodes $i$ and $j$ might not satisfy the triangle inequality, which represents a general real-world case of the problem formulation. 
Unlike the~\textit{akb} set, $\mu_1=300$ and $\mu_2=0.014$ in the~\textit{jd} set are directly given based on the estimated values in the application.

\subsubsection{Parameter Setting}
We determined the parameter settings of HMA through preliminary experiments.
The settings for different problem scales are detailed in Table~\ref{tab: parameter_setting}.
Specifically, HMA requires setting seven parameters: $G_1$, $G_2$, $N$, $\alpha$, $B$, $sr$, and $w_1, w_2$.
$G_1$ and $G_2$ denote the longest consecutive iterations/generations without improvement in LNS and MS of HMA, respectively, while $N$ represents the population size in MS (see Algorithm~\ref{alg: Hybrid Search Framework}).
$\alpha$ and $B$ refer to the proportionality parameter and the maximum number of generations in PSI, respectively.
$sr$ specifies the selection range in station ranking list (see Algorithm~\ref{alg: Parallel-Sequential Station Insertion}). 
$w_1$ and $w_2$ are the lower and upper bounds for the proportion of customers removed in the destroy-repair operator (see Algorithm~\ref{alg: Hybrid Search Framework}).
All the experiments were run on an Intel Xeon Gold 6240 machine with 377 GB RAM and 32 cores (2.60 GHz, 25 MB Cache), running Rocky Linux 8.5.
HMA was implemented in C++. 

\subsection{Results on Small-scale and Medium-scale Instances}
We compared HMA with two recently proposed state-of-the-art algorithms, Adapt-CMSA-STD and Adapt-CMSA-SETCOV, on the~\textit{akb} set.
Since their source code is not publicly available, we obtained the testing results directly from the original publications~\cite{akbay2022application,akbay2023cmsa}. 
To ensure a fair comparison, we rescaled the computation time to a common measure considering the CPUs used. 
Specifically, we rescaled the time based on the \href{www.passmark.com}{PassMark score} of the CPUs used in the studies relative to our 2.60~GHz Intel Xeon Gold 6240. 
The PassMark score, which reflects the performance of a single CPU core, is 1398 for the 2.93 GHz X5670 used in\cite{akbay2022application,akbay2023cmsa}, whereas our CPU scores 1990. The rescaled times (in seconds) are reported in Tables~\ref{tab: small-scale} and~\ref{tab: medium-scale}.
% The PassMark score, as the performance of a single core of the respective CPU, for the 2.93~GHz X5670 used in~\cite{akbay2022application,akbay2023cmsa} is 1398, while the score of our CPU is 1990. 
% The rescaled time in seconds is given in Tables~\ref{tab: small-scale} and~\ref{tab: medium-scale}. 
% Additionally, the run time limit on the small-scale and medium-scale instances is rescaled to 105 and 630 seconds.
Additionally, after rescaling, the run time limits for the small-scale and medium-scale instances were set to 105 and 630 seconds, respectively.

Tables~\ref{tab: small-scale} and \ref{tab: medium-scale} are structured as follows. 
The first column lists the instance names. 
The ``m'' columns indicate the number of EVs used in the respective solutions. 
For Adapt-CMSA-STD, Adapt-CMSA-SETCOV, and HMA, the ``best'' column refers to the $TC$ of the best solution found in 10 independent runs, while the ``avg'' column refers to the average $TC$ of the 10 runs.
The ``time'' column shows the computation time (in seconds) for CPLEX and the average computation time for Adapt-CMSA-STD, Adapt-CMSA-SETCOV, and HMA to find the best solutions in each run, respectively. 
Finally, the ``gap'' column shows the relative improvement of the best solution obtained by HMA compared to the existing best-known solution. Specifically, $gap=[TC_{\text{HMA}} - TC_{\text{best known}}]/TC_{\text{best known}}$. A negative value of $gap$ indicates that HMA has achieved an improvement in solution quality.
For each instance, the best ``best'' value and the best ``avg'' value are highlighted in gray.
% For each instance, the best performance w.r.t. ``best'' and ``avg'' are highlighted in gray, respectively.

Based on the numerical results in Table~\ref{tab: small-scale}, the following observations can be made. 
On small-scale instances, the three metaheuristic algorithms were able to obtain optimal solutions in much shorter computation time compared to CPLEX, and even found better solutions than CPLEX in instances \textit{r202C15} and \textit{rc204C15}. 
Overall, Adapt-CMSA-STD, Adapt-CMSA-SETCOV, and HMA exhibit comparable performance, with no statistically significant differences in their best or average solution quality, likely due to the small solution space of these instances (see Section~\ref{subsec: statistical test} for the statistical test).
The \(gap\) value is \(0.00\%\) for nearly all instances except \textit{r202C15}, indicating that HMA achieved solution quality equivalent to the best-known solutions in most cases.
Furthermore, HMA demonstrates minimal variation in computation time as the problem scale increases, with an average runtime of 2.20 seconds, which is much shorter than that of the other two metaheuristic algorithms.
% On small-scale instances, compared to CPLEX, the three metaheuristic algorithms found optimal solutions in a much shorter time, even found better solutions than CPLEX in instance $r202C15$ and $r204C15$. 
% Overall, Adapt-CMSA-STD, Adapt-CMSA-SETCOV, and HMA are comparable, showing no significant difference in their best or average values due to the small solution space~(see Section~\ref{subsec: statistical test} for the statistical test). 
% {\color{blue}The \(gap\) value is \(0.00\%\) in almost all instances except instance $r202C15$, indicating that the solution quality achieved by HMA is equivalent to the best-known solution.}
% Moreover, HMA shows minimal variation in computation time as the problem scale increases, with an average time of 2.20 seconds, which is much shorter than that of the other three algorithms.
% Moreover, it is worth mentioning that HMA exhibits minimal computation time variation as the problem scale increases, with an average computation time of 2.20 seconds, which is much shorter than that of the other three algorithms.

Generally, the performance of algorithms on instances of larger scales is of more interest. 
Overall, HMA is the best-performing algorithm on medium-scale instances as shown in Table~\ref{tab: medium-scale}. 
In the \textit{r$\sim$} and~\textit{rc$\sim$} subsets, the best and average solution quality obtained by HMA are consistently better than those of Adapt-CMSA-STD and Adapt-CMSA-SETCOV. 
Moreover, in the \textit{c$\sim$} subsets, HMA outperformed the other two algorithms in all except four instances.
Note that for the \textit{akb} set, minimizing the number of used EVs has a higher priority than minimizing $TD$, as indicated by the settings $\mu_1 = 1000$ and $\mu_2 = 1$ in Eq.~\eqref{eq:5}.
It is worth mentioning that HMA obtained competitive solutions with fewer EVs in 23 out of 56 medium-scale instances.
More specifically, these instances include 4 out of 9 in the \textit{c1$\sim$} subset, 7 out of 12 in the \textit{r1$\sim$} subset, 4 out of 11 in the \textit{r2$\sim$} subset, 6 out of 8 in the \textit{rc1$\sim$} subset, and 2 out of 8 in the \textit{rc2$\sim$} subset.
It is observed that in all the aforementioned instances, the $gap$ values are below $-5.00\%$, indicating a significant improvement. 
For the remaining instances where the same number of EVs is used, the $gap$ values are nearly $0.00\%$, yet remain negative, which suggests that HMA still achieved solution quality better than the best-known solutions in those cases.
Considering that customer distribution and constraint intensity vary across different instance types~(see Section~\ref{benchmark}), such results indicate that HMA performed both strongly and robustly.
% In instances 
% c102, c104, c106, c107, 
% r101, r103, r104, r106, r109, r110, r112, r201, r204, r207, r208, 
% rc101, rc102, rc103, rc104, rc107, rc108, rc203, rc206, 
Moreover, HMA required considerably less computation time, using only about 60\% of the time consumed by the other two algorithms.

Compared to Adapt-CMSA-STD and Adapt-CMSA-SETCOV, HMA exhibits a remarkable performance advantage on medium-scale instances. 
This improvement is primarily due to the fact that, unlike Adapt-CMSA-STD and Adapt-CMSA-SETCOV, which maintain and refine only a single elite solution throughout the search process, HMA employs a memetic search that initializes the population based on the elite solution obtained from the large neighborhood search.
This strategy enables effective exploitation of the high-quality information contained in the elite solution, while maintaining population diversity to reduce the risk of premature convergence. 
In addition, the combined use of PSI and SSI, helps avoid getting trapped in poor-quality local optima.
% As a result, HMA is capable of finding high-quality solutions on medium-scale instances.
The effectiveness of these novel components integrated into HMA will be further analyzed in Section~\ref{subsec: Effectiveness of Each Component in HMA}.

In summary, the observations presented above collectively indicate that HMA outperforms recent state-of-the-art algorithms across a broad range of problem instances. 
Notably, it has significantly raised the performance bar on the \textit{akb} set.

\begin{table*}[]
\caption{Comparative results on medium-scale instances in the \( akb \) set.}
\label{tab: medium-scale}
\resizebox{\textwidth}{!}{%
\begin{tabular}{ccccclcccclccccc}
\toprule
 & \multicolumn{4}{c}{Adapt-CMSA-STD} &  & \multicolumn{4}{c}{Adapt-CMSA-SETCOV} &  & \multicolumn{5}{c}{HMA}  \\ \cline{2-5} \cline{7-10} \cline{12-16}
\multirow{-2}{*}{Instance name} & m & best & avg & time &  & m & best & avg & time &  & m & best & avg & time & gap \\ \midrule
c101 & 12 & 13043.40 & \cellcolor[HTML]{D9D9D9}13043.42 & 270.56 &  & 12 & 13057.80 & 13063.54 & 205.53 &  & 12 & \cellcolor[HTML]{D9D9D9}13043.38 & 13043.43 & 12.57 & 0.00\% \\
c102 & 11 & 12056.80 & 12920.23 & 393.95 &  & 11 & 12073.10 & 12944.34 & 329.34 &  & 10 & \cellcolor[HTML]{D9D9D9}11089.80 & \cellcolor[HTML]{D9D9D9}12033.33 & 83.87 & -8.02\% \\
c103 & 11 & 12004.70 & 12026.90 & 317.63 &  & 10 & 11134.90 & 11917.80 & 504.47 &  & 10 & \cellcolor[HTML]{D9D9D9}10945.45 & \cellcolor[HTML]{D9D9D9}10945.48 & 136.32 & -1.70\% \\
c104 & 10 & 10872.80 & 11353.78 & 442.55 &  & 10 & 10870.70 & 10876.49 & 427.43 &  & 9 & \cellcolor[HTML]{D9D9D9}9890.98 & \cellcolor[HTML]{D9D9D9}10767.31 & 231.74 & -9.01\% \\
c105 & 11 & \cellcolor[HTML]{D9D9D9}12023.80 & 12341.60 & 394.88 &  & 11 & 12034.10 & \cellcolor[HTML]{D9D9D9}12068.86 & 409.38 &  & 11 & 12030.81 & 12130.20 & 64.75 & 0.06\% \\
c106 & 11 & 12013.10 & 12438.06 & 458.04 &  & 11 & 12025.70 & 12059.29 & 305.45 &  & 10 & \cellcolor[HTML]{D9D9D9}11040.32 & \cellcolor[HTML]{D9D9D9}11825.37 & 73.50 & -8.10\% \\
c107 & 11 & 12006.40 & 12023.97 & 378.24 &  & 11 & 12026.70 & 12046.38 & 276.09 &  & 10 & \cellcolor[HTML]{D9D9D9}11021.16 & \cellcolor[HTML]{D9D9D9}11815.55 & 67.93 & -8.21\% \\
c108 & 11 & 11994.70 & 12016.10 & 407.11 &  & 10 & 11025.80 & 11822.60 & 391.00 &  & 10 & \cellcolor[HTML]{D9D9D9}11006.10 & \cellcolor[HTML]{D9D9D9}11597.49 & 137.57 & -0.18\% \\
c109 & 10 & 11042.20 & 11885.30 & 502.22 &  & 10 & 10941.00 & 11180.77 & 524.19 &  & 10 & \cellcolor[HTML]{D9D9D9}10911.47 & \cellcolor[HTML]{D9D9D9}10936.41 & 184.63 & -0.27\% \\
c201 & 4 & \cellcolor[HTML]{D9D9D9}4629.95 & \cellcolor[HTML]{D9D9D9}4629.95 & 26.41 &  & 4 & 4678.37 & 4703.43 & 274.65 &  & 4 & \cellcolor[HTML]{D9D9D9}4629.95 & \cellcolor[HTML]{D9D9D9}4629.95 & 9.55 & 0.00\% \\
c202 & 4 & \cellcolor[HTML]{D9D9D9}4629.95 & \cellcolor[HTML]{D9D9D9}4629.95 & 192.19 &  & 4 & 4664.26 & 4706.94 & 277.38 &  & 4 & \cellcolor[HTML]{D9D9D9}4629.95 & \cellcolor[HTML]{D9D9D9}4629.95 & 34.02 & 0.00\% \\
c203 & 4 & 4632.27 & 4690.06 & 520.20 &  & 4 & 4641.45 & 4734.31 & 349.57 &  & 4 & \cellcolor[HTML]{D9D9D9}4629.95 & \cellcolor[HTML]{D9D9D9}4629.95 & 47.47 & -0.05\% \\
c204 & 4 & 4633.08 & 4665.78 & 563.25 &  & 4 & 4660.64 & 4737.07 & 503.66 &  & 4 & \cellcolor[HTML]{D9D9D9}4628.91 & \cellcolor[HTML]{D9D9D9}4628.91 & 102.73 & -0.09\% \\
c205 & 4 & \cellcolor[HTML]{D9D9D9}4629.95 & \cellcolor[HTML]{D9D9D9}4629.95 & 54.00 &  & 4 & \cellcolor[HTML]{D9D9D9}4629.95 & \cellcolor[HTML]{D9D9D9}4629.95 & 88.12 &  & 4 & \cellcolor[HTML]{D9D9D9}4629.95 & \cellcolor[HTML]{D9D9D9}4629.95 & 31.58 & 0.00\% \\
c206 & 4 & \cellcolor[HTML]{D9D9D9}4629.95 & \cellcolor[HTML]{D9D9D9}4629.95 & 149.83 &  & 4 & \cellcolor[HTML]{D9D9D9}4629.95 & \cellcolor[HTML]{D9D9D9}4629.95 & 142.64 &  & 4 & \cellcolor[HTML]{D9D9D9}4629.95 & 4630.26 & 39.25 & 0.00\% \\
c207 & 4 & \cellcolor[HTML]{D9D9D9}4629.95 & \cellcolor[HTML]{D9D9D9}4629.95 & 179.74 &  & 4 & \cellcolor[HTML]{D9D9D9}4629.95 & 4635.27 & 182.89 &  & 4 & \cellcolor[HTML]{D9D9D9}4629.95 & 4630.89 & 62.34 & 0.00\% \\
c208 & 4 & \cellcolor[HTML]{D9D9D9}4629.95 & \cellcolor[HTML]{D9D9D9}4629.95 & 200.06 &  & 4 & \cellcolor[HTML]{D9D9D9}4629.95 & \cellcolor[HTML]{D9D9D9}4629.95 & 183.86 &  & 4 & \cellcolor[HTML]{D9D9D9}4629.95 & \cellcolor[HTML]{D9D9D9}4629.95 & 50.89 & 0.00\% \\
r101 & 18 & 19633.80 & 19939.79 & 459.20 &  & 18 & 19640.60 & 19661.15 & 476.75 &  & 17 & \cellcolor[HTML]{D9D9D9}18636.96 & \cellcolor[HTML]{D9D9D9}19512.43 & 75.46 & -5.08\% \\
r102 & 17 & 18470.80 & 19292.16 & 496.96 &  & 16 & 17474.10 & 17696.35 & 561.10 &  & 16 & \cellcolor[HTML]{D9D9D9}17424.75 & \cellcolor[HTML]{D9D9D9}17526.29 & 115.80 & -0.28\% \\
r103 & 15 & 16296.50 & 17050.75 & 499.65 &  & 14 & 15280.30 & 15306.17 & 449.49 &  & 13 & \cellcolor[HTML]{D9D9D9}14209.21 & \cellcolor[HTML]{D9D9D9}14721.75 & 187.46 & -7.01\% \\
r104 & 13 & 14141.10 & 14255.53 & 433.34 &  & 12 & 13084.30 & 13111.31 & 538.23 &  & 11 & \cellcolor[HTML]{D9D9D9}12051.19 & \cellcolor[HTML]{D9D9D9}12066.53 & 285.75 & -7.90\% \\
r105 & 15 & 16389.20 & 17212.83 & 477.88 &  & 14 & 15471.30 & 16346.10 & 429.92 &  & 14 & \cellcolor[HTML]{D9D9D9}15366.33 & \cellcolor[HTML]{D9D9D9}16159.23 & 146.98 & -0.68\% \\
r106 & 15 & 16292.00 & 16836.67 & 492.99 &  & 14 & 15314.80 & 15441.68 & 524.23 &  & 13 & \cellcolor[HTML]{D9D9D9}14265.55 & \cellcolor[HTML]{D9D9D9}14373.31 & 222.14 & -6.85\% \\
r107 & 13 & 14168.90 & 15016.67 & 478.40 &  & 12 & 13140.10 & 13669.50 & 550.64 &  & 12 & \cellcolor[HTML]{D9D9D9}13112.66 & \cellcolor[HTML]{D9D9D9}13118.04 & 226.39 & -0.21\% \\
r108 & 12 & 13079.80 & 13531.30 & 468.60 &  & 11 & 12073.70 & 12998.17 & 521.61 &  & 11 & \cellcolor[HTML]{D9D9D9}12016.74 & \cellcolor[HTML]{D9D9D9}12025.10 & 289.00 & -0.47\% \\
r109 & 14 & 15237.30 & 15674.51 & 533.66 &  & 13 & 14220.80 & 14468.12 & 523.07 &  & 12 & \cellcolor[HTML]{D9D9D9}13203.74 & \cellcolor[HTML]{D9D9D9}13993.94 & 237.95 & -7.15\% \\
r110 & 13 & 14170.20 & 14905.73 & 370.99 &  & 12 & 13114.30 & 13544.76 & 541.43 &  & 11 & \cellcolor[HTML]{D9D9D9}12070.04 & \cellcolor[HTML]{D9D9D9}12482.82 & 270.95 & -7.96\% \\
r111 & 12 & 13144.20 & 14584.19 & 489.45 &  & 12 & 13148.80 & 13965.98 & 503.08 &  & 12 & \cellcolor[HTML]{D9D9D9}13070.64 & \cellcolor[HTML]{D9D9D9}13077.15 & 225.63 & -0.56\% \\
r112 & 12 & 13155.60 & 14053.56 & 331.29 &  & 12 & 13044.10 & 13078.65 & 597.35 &  & 11 & \cellcolor[HTML]{D9D9D9}12001.79 & \cellcolor[HTML]{D9D9D9}12012.62 & 233.83 & -7.99\% \\
r201 & 4 & 5192.33 & 5216.92 & 506.12 &  & 4 & 5276.75 & 5363.04 & 131.64 &  & 3 & \cellcolor[HTML]{D9D9D9}4268.88 & \cellcolor[HTML]{D9D9D9}4985.55 & 264.02 & -17.78\% \\
r202 & 3 & 4250.70 & 5020.88 & 483.79 &  & 3 & 4193.33 & 4940.22 & 615.58 &  & 3 & \cellcolor[HTML]{D9D9D9}4054.75 & \cellcolor[HTML]{D9D9D9}4063.20 & 363.02 & -3.30\% \\
r203 & 3 & 3942.74 & 4352.52 & 609.82 &  & 3 & 3985.02 & 4060.18 & 577.82 &  & 3 & \cellcolor[HTML]{D9D9D9}3898.68 & \cellcolor[HTML]{D9D9D9}3910.77 & 541.38 & -1.12\% \\
r204 & 3 & 3820.72 & 3854.31 & 553.01 &  & 3 & 3793.76 & 3827.81 & 615.67 &  & 2 & \cellcolor[HTML]{D9D9D9}2815.58 & \cellcolor[HTML]{D9D9D9}3544.13 & 438.26 & -25.78\% \\
r205 & 3 & 4055.28 & 4124.64 & 514.20 &  & 3 & 4065.06 & 4126.45 & 253.23 &  & 3 & \cellcolor[HTML]{D9D9D9}3995.35 & \cellcolor[HTML]{D9D9D9}4003.09 & 512.99 & -1.48\% \\
r206 & 3 & 3978.10 & 4065.05 & 531.38 &  & 3 & 3991.44 & 4047.16 & 550.82 &  & 3 & \cellcolor[HTML]{D9D9D9}3925.95 & \cellcolor[HTML]{D9D9D9}3938.12 & 600.35 & -1.31\% \\
r207 & 3 & 3878.91 & 3910.07 & 463.55 &  & 3 & 3881.97 & 3918.29 & 616.95 &  & 2 & \cellcolor[HTML]{D9D9D9}2851.75 & \cellcolor[HTML]{D9D9D9}3534.37 & 494.20 & -26.48\% \\
r208 & 3 & 3791.27 & 3829.39 & 596.95 &  & 3 & 3732.80 & 3776.20 & 629.30 &  & 2 & \cellcolor[HTML]{D9D9D9}2740.06 & \cellcolor[HTML]{D9D9D9}2756.80 & 589.42 & -26.60\% \\
r209 & 3 & 3975.64 & 4015.90 & 467.75 &  & 3 & 3933.55 & 3977.24 & 538.01 &  & 3 & \cellcolor[HTML]{D9D9D9}3877.24 & \cellcolor[HTML]{D9D9D9}3885.71 & 636.14 & -1.43\% \\
r210 & 3 & 3920.37 & 3984.78 & 530.97 &  & 3 & 3926.79 & 3961.42 & 520.11 &  & 3 & \cellcolor[HTML]{D9D9D9}3846.62 & \cellcolor[HTML]{D9D9D9}3854.33 & 555.78 & -1.88\% \\
r211 & 3 & 3814.42 & 3893.38 & 580.24 &  & 3 & 3824.47 & 3857.62 & 561.62 &  & 3 & \cellcolor[HTML]{D9D9D9}3772.67 & \cellcolor[HTML]{D9D9D9}3789.62 & 477.61 & -1.09\% \\
rc101 & 16 & 17667.70 & 18513.67 & 504.48 &  & 16 & 17696.20 & 17741.63 & 442.05 &  & 15 & \cellcolor[HTML]{D9D9D9}16661.53 & \cellcolor[HTML]{D9D9D9}17462.90 & 152.22 & -5.69\% \\
rc102 & 16 & 17576.80 & 17909.78 & 392.20 &  & 15 & 16558.20 & 16628.46 & 422.70 &  & 14 & \cellcolor[HTML]{D9D9D9}15510.31 & \cellcolor[HTML]{D9D9D9}15618.29 & 256.61 & -6.33\% \\
rc103 & 14 & 15366.90 & 16245.06 & 537.04 &  & 13 & 14358.20 & 14999.16 & 485.96 &  & 12 & \cellcolor[HTML]{D9D9D9}13343.56 & \cellcolor[HTML]{D9D9D9}13648.08 & 297.57 & -7.07\% \\
rc104 & 13 & 14270.50 & 14315.17 & 447.54 &  & 12 & 13222.50 & 13261.65 & 490.66 &  & 11 & \cellcolor[HTML]{D9D9D9}12150.47 & \cellcolor[HTML]{D9D9D9}12168.35 & 388.45 & -8.11\% \\
rc105 & 15 & 16500.90 & 16933.42 & 458.25 &  & 14 & 15470.70 & 15820.90 & 449.50 &  & 14 & \cellcolor[HTML]{D9D9D9}15447.52 & \cellcolor[HTML]{D9D9D9}15454.71 & 164.64 & -0.15\% \\
rc106 & 14 & 15432.50 & 16069.05 & 444.10 &  & 13 & 14448.30 & 15249.17 & 406.17 &  & 13 & \cellcolor[HTML]{D9D9D9}14388.12 & \cellcolor[HTML]{D9D9D9}14609.85 & 190.86 & -0.42\% \\
rc107 & 13 & 14313.40 & 14437.31 & 540.67 &  & 12 & 13276.50 & 13386.51 & 518.89 &  & 11 & \cellcolor[HTML]{D9D9D9}12234.76 & \cellcolor[HTML]{D9D9D9}13048.02 & 232.42 & -7.85\% \\
rc108 & 12 & 13226.00 & 13891.71 & 435.95 &  & 12 & 13184.70 & 13214.50 & 504.05 &  & 11 & \cellcolor[HTML]{D9D9D9}12140.28 & \cellcolor[HTML]{D9D9D9}12152.42 & 271.54 & -7.92\% \\
rc201 & 4 & 5504.77 & 5819.06 & 494.52 &  & 4 & 5617.75 & 5786.48 & 226.81 &  & 4 & \cellcolor[HTML]{D9D9D9}5444.90 & \cellcolor[HTML]{D9D9D9}5450.77 & 254.36 & -1.09\% \\
rc202 & 4 & 5324.64 & 5442.41 & 416.84 &  & 4 & 5436.63 & 5541.80 & 138.34 &  & 4 & \cellcolor[HTML]{D9D9D9}5240.40 & \cellcolor[HTML]{D9D9D9}5249.87 & 192.66 & -1.58\% \\
rc203 & 4 & 5109.88 & 5177.69 & 452.57 &  & 4 & 5086.44 & 5159.15 & 531.94 &  & 3 & \cellcolor[HTML]{D9D9D9}4078.55 & \cellcolor[HTML]{D9D9D9}4100.32 & 597.82 & -19.82\% \\
rc204 & 3 & 4036.49 & 4525.03 & 523.68 &  & 3 & 3962.88 & 4252.28 & 631.65 &  & 3 & \cellcolor[HTML]{D9D9D9}3892.97 & \cellcolor[HTML]{D9D9D9}3902.06 & 528.96 & -1.76\% \\
rc205 & 4 & 5260.14 & 5338.50 & 426.74 &  & 4 & 5285.41 & 5375.54 & 221.24 &  & 4 & \cellcolor[HTML]{D9D9D9}5149.13 & \cellcolor[HTML]{D9D9D9}5152.79 & 368.18 & -2.11\% \\
rc206 & 4 & 5234.55 & 5289.90 & 470.82 &  & 4 & 5210.57 & 5275.72 & 395.42 &  & 3 & \cellcolor[HTML]{D9D9D9}4200.69 & \cellcolor[HTML]{D9D9D9}4323.33 & 517.29 & -19.38\% \\
rc207 & 3 & 4150.60 & 4930.81 & 488.10 &  & 3 & 4197.81 & 4650.01 & 607.24 &  & 3 & \cellcolor[HTML]{D9D9D9}3979.65 & \cellcolor[HTML]{D9D9D9}4019.57 & 490.17 & -4.12\% \\
rc208 & 3 & 3977.50 & 4046.09 & 535.72 &  & 3 & 3920.17 & 4002.79 & 527.17 &  & 3 & \cellcolor[HTML]{D9D9D9}3838.78 & \cellcolor[HTML]{D9D9D9}3857.49 & 457.65 & -2.08\% \\ \midrule
avg & 8.48 & 9568.86 & 9905.20 & 435.54 &  & 8.18 & 9266.06 & 9479.93 & 432.20 &  & 7.77 & 8807.09 & 9030.14 & 262.90 & -5.20\% \\
\bottomrule
\end{tabular}%
}
\end{table*}
\begin{table*}[t]
\caption{Comparative results on large-scale instances in the \( jd \) set with a short time limit.}
\label{tab: large-scale short-time-limit}
\resizebox{\textwidth}{!}{%
\begin{tabular}{ccccclcccclcccclcccc}
\toprule
 & \multicolumn{4}{c}{Adapt-CMSA-STD} &  & \multicolumn{4}{c}{Adapt-CMSA-SETCOV} &  & \multicolumn{4}{c}{HMA w/o BCD} &  & \multicolumn{3}{c}{HMA} &  \\ \cline{2-5} \cline{7-10} \cline{12-15} \cline{17-19}
\multirow{-2}{*}{Instance name} & best & avg$\pm$std & time & $\Delta\%$&  & best & avg$\pm$std & time & $\Delta\%$ &  & best & avg$\pm$std & time & $\Delta\%$ &  & best & avg$\pm$std & time & \multirow{-2}{*}{p-value} \\
\midrule
jd200\_1 & 74060.10 & 75718.31$\pm$908.58 & 1782.33 &  &  & 76125.24 & 78311.24$\pm$1416.76 & 1803.16 &  &  & 71043.42 & 72243.96$\pm$614.67 & 1884.42 &  &  & \cellcolor[HTML]{D9D9D9}70979.75 & 72348.31$\pm$724.19 & 1821.75 & 0.6501 \\
jd200\_2 & 66255.36 & 67505.53$\pm$655.09 & 1797.81 &  &  & 71075.99 & 71965.57$\pm$704.99 & 1801.42 &  &  & 59660.84 & 60785.77$\pm$551.51 & 1892.70 &  &  & \cellcolor[HTML]{D9D9D9}59460.27 & 60688.60$\pm$697.48 & 1825.64 & 0.7624 \\
jd200\_3 & 68219.71 & 69710.26$\pm$763.65 & 1686.48 &  &  & 70656.63 & 71918.09$\pm$792.00 & 1803.03 &  &  & 66634.23 & 67043.56$\pm$269.90 & 1872.56 &  &  & \cellcolor[HTML]{D9D9D9}66254.27 & 66999.61$\pm$526.33 & 1839.70 & 0.8798 \\
jd200\_4 & 72557.01 & 74701.11$\pm$1482.53 & 1800.69 & \multirow{-4}{*}{9.45\%} &  & 81696.93 & 84613.90$\pm$2035.06 & 1802.73 & \multirow{-4}{*}{16.87\%} &  & 60988.30 & 62033.06$\pm$488.03 & 1873.78 & \multirow{-4}{*}{0.27\%} &  & \cellcolor[HTML]{D9D9D9}60941.60 & 61753.23$\pm$523.17 & 1843.03 & 0.2568 \\
jd400\_1 & 141561.75 & 142694.05$\pm$1128.17 & 5406.36 &  &  & 143045.76 & 145872.43$\pm$1772.19 & 5406.09 &  &  & 116029.37 & 117199.07$\pm$954.98 & 5784.07 &  &  & \cellcolor[HTML]{D9D9D9}115981.66 & 116987.12$\pm$952.18 & 5443.33 & 0.4963 \\
jd400\_2 & 136842.18 & 139007.05$\pm$1738.07 & 5421.45 &  &  & 137495.23 & 141242.66$\pm$2324.07 & 5404.31 &  &  & 116527.33 & 117584.91$\pm$736.78 & 5734.43 &  &  & \cellcolor[HTML]{D9D9D9}116205.05 & 117098.09$\pm$563.09 & 5430.44 & 0.1041 \\
jd400\_3 & 118665.20 & 120745.11$\pm$1413.70 & 5410.98 &  &  & 120888.17 & 122891.09$\pm$1530.63 & 5405.72 &  &  & 109030.69 & 109934.10$\pm$508.53 & 5576.75 &  &  & \cellcolor[HTML]{D9D9D9}108441.53 & \textbf{109140.65$\pm$451.85} & 5421.66 & 0.0032 \\
jd400\_4 & 137312.31 & 141331.04$\pm$2085.47 & 5404.06 & \multirow{-4}{*}{13.94\%} &  & 137863.84 & 142086.80$\pm$1877.16 & 5406.27 & \multirow{-4}{*}{15.02\%} &  & 128988.11 & 129937.62$\pm$710.96 & 5761.35 & \multirow{-4}{*}{0.23\%} &  & \cellcolor[HTML]{D9D9D9}128894.04 & 129654.11$\pm$858.13 & 5428.58 & 0.1509 \\
jd600\_1 & 213824.87 & 221616.09$\pm$7471.96 & 9097.22 &  &  & 210722.03 & 215743.58$\pm$2767.03 & 8988.48 &  &  & 167369.71 & 169339.19$\pm$1098.95 & 9351.29 &  &  & \cellcolor[HTML]{D9D9D9}167297.70 & 168836.18$\pm$1087.42 & 9033.06 & 0.3258 \\
jd600\_2 & 206849.40 & 215497.97$\pm$6446.15 & 9103.28 &  &  & 205943.56 & 210498.81$\pm$2071.06 & 9008.39 &  &  & 171996.76 & 174044.48$\pm$1610.00 & 9643.47 &  &  & \cellcolor[HTML]{D9D9D9}171512.97 & 173480.89$\pm$1384.65 & 9019.82 & 0.3447 \\
jd600\_3 & 173412.99 & 176186.74$\pm$1944.15 & 9243.68 &  &  & 172977.87 & 175893.80$\pm$1829.45 & 8887.64 &  &  & 154336.98 & 155381.50$\pm$773.48 & 9610.60 &  &  & \cellcolor[HTML]{D9D9D9}153092.64 & \textbf{154325.16$\pm$840.34} & 9037.23 & 0.0191 \\
jd600\_4 & 201123.62 & 211620.07$\pm$7977.66 & 9009.64 & \multirow{-4}{*}{23.91\%} &  & 197929.76 & 201306.90$\pm$2408.88 & 9008.28 & \multirow{-4}{*}{22.72\%} &  & 150513.45 & 151553.99$\pm$649.75 & 9753.66 & \multirow{-4}{*}{0.35\%} &  & \cellcolor[HTML]{D9D9D9}150123.89 & \textbf{150975.72$\pm$609.08} & 9049.35 & 0.0343 \\
jd800\_1 & 275924.88 & 313298.97$\pm$20935.22 & 12617.93 &  &  & 271098.99 & 276999.84$\pm$4688.42 & 12557.60 &  &  & 216364.67 & 217948.77$\pm$1027.41 & 13710.86 &  &  & \cellcolor[HTML]{D9D9D9}214710.04 & 216810.78$\pm$1542.74 & 12625.32 & 0.0757 \\
jd800\_2 & 269445.25 & 312947.07$\pm$22210.55 & 12667.52 &  &  & 268047.17 & 275025.17$\pm$8658.96 & 12607.02 &  &  & 226218.76 & 228162.86$\pm$1987.21 & 13523.34 &  &  & \cellcolor[HTML]{D9D9D9}223610.37 & 226527.29$\pm$2364.16 & 12619.90 & 0.0963 \\
jd800\_3 & 234915.71 & 257492.90$\pm$17983.73 & 13053.00 &  &  & 232052.94 & 239688.51$\pm$9680.09 & 12611.89 &  &  & 206772.88 & 208548.00$\pm$1067.70 & 13401.35 &  &  & \cellcolor[HTML]{D9D9D9}205215.92 & \textbf{207375.35$\pm$1432.89} & 12619.30 & 0.0494 \\
jd800\_4 & 282966.34 & 297859.86$\pm$13680.91 & 12613.94 & \multirow{-4}{*}{27.33\%} &  & 260642.67 & 266243.72$\pm$4288.39 & 12610.64 & \multirow{-4}{*}{23.39\%} &  & 196076.58 & 197143.07$\pm$637.84 & 13603.29 & \multirow{-4}{*}{0.94\%} &  & \cellcolor[HTML]{D9D9D9}194004.95 & 196257.28$\pm$1325.29 & 12621.05 & 0.0696 \\
jd1000\_1 & 378549.15 & 409496.27$\pm$20008.55 & 16265.94 &  &  & 327562.85 & 345987.18$\pm$20488.34 & 16214.39 &  &  & 261338.90 & 262810.90$\pm$1082.93 & 17483.69 &  &  & \cellcolor[HTML]{D9D9D9}257410.81 & \textbf{260877.67$\pm$2210.64} & 16224.57 & 0.0343 \\
jd1000\_2 & 369338.23 & 408955.96$\pm$26354.10 & 16232.57 &  &  & 335504.67 & 351073.09$\pm$20290.60 & 16214.16 &  &  & 276597.69 & 279318.78$\pm$1341.59 & 16643.43 &  &  & \cellcolor[HTML]{D9D9D9}274079.42 & \textbf{276515.30$\pm$1881.23} & 16224.80 & 0.0028 \\
jd1000\_3 & 314790.77 & 361696.11$\pm$32886.34 & 16233.94 &  &  & 286785.82 & 304623.25$\pm$21348.50 & 16218.22 &  &  & 256229.79 & 258040.35$\pm$1467.13 & 16631.34 &  &  & \cellcolor[HTML]{D9D9D9}252772.63 & \textbf{256102.57$\pm$2369.95} & 16221.87 & 0.0494 \\
jd1000\_4 & 349809.78 & 391564.48$\pm$26270.35 & 16229.30 & \multirow{-4}{*}{37.94\%} &  & 320782.78 & 333245.22$\pm$15639.71 & 16221.33 & \multirow{-4}{*}{24.12\%} &  & 243354.91 & 244808.00$\pm$1235.27 & 17552.66 & \multirow{-4}{*}{1.24\%} &  & \cellcolor[HTML]{D9D9D9}240559.58 & \textbf{242952.46$\pm$2044.65} & 16218.06 & 0.0284 \\
\bottomrule
\end{tabular}%
}
\end{table*}
\begin{table*}[t]
\caption{Comparative results on large-scale instances in the \( jd \) set with a long time limit.}
\label{tab: large-scale long-time-limit}
\resizebox{\textwidth}{!}{%
\begin{tabular}{ccccclcccclcccclcccc}
\toprule
 & \multicolumn{4}{c}{Adapt-CMSA-STD} &  & \multicolumn{4}{c}{Adapt-CMSA-SETCOV} &  & \multicolumn{4}{c}{HMA w/o BCD} &  & \multicolumn{3}{c}{HMA} &  \\ \cline{2-5} \cline{7-10} \cline{12-15} \cline{17-19}
\multirow{-2}{*}{Instance name} & best & avg$\pm$std & time & $\Delta\%$&  & best & avg$\pm$std & time & $\Delta\%$ &  & best & avg$\pm$std & time & $\Delta\%$ &  & best & avg$\pm$std & time & \multirow{-2}{*}{p-value} \\
\midrule
jd200\_1 & 73644.20 & 74971.07$\pm$892.62 & 3601.47 &  &  & 76125.24 & 78512.42$\pm$1482.09 & 3609.07 &  &  & 70740.90 & 71335.98$\pm$403.87 & 3678.82 &  &  & \cellcolor[HTML]{D9D9D9}70440.57 & 71409.68$\pm$512.27 & 3510.36 & 0.6501 \\
jd200\_2 & 65007.91 & 66759.93$\pm$1100.01 & 3577.74 &  &  & 71075.99 & 71940.92$\pm$728.62 & 3605.35 &  &  & 59389.70 & 60004.86$\pm$419.68 & 3661.48 &  &  & \cellcolor[HTML]{D9D9D9}58530.22 & 59823.10$\pm$570.84 & 3612.10 & 0.3643 \\
jd200\_3 & 67644.15 & 69089.36$\pm$946.59 & 3196.51 &  &  & 71600.11 & 72482.85$\pm$646.16 & 3604.60 &  &  & \cellcolor[HTML]{D9D9D9}65201.82 & 65844.88$\pm$445.08 & 3544.28 &  &  & 65812.71 & 66206.45$\pm$284.52 & 3534.71 & 0.0588 \\
jd200\_4 & 71929.75 & 74205.32$\pm$1681.07 & 3601.88 & \multirow{-4}{*}{9.17\%} &  & 81696.93 & 84631.25$\pm$2017.07 & 3609.60 & \multirow{-4}{*}{18.16\%} &  & \cellcolor[HTML]{D9D9D9}60747.28 & 60983.75$\pm$250.76 & 3571.75 & \multirow{-4}{*}{0.21\%} &  & 60815.01 & 61008.17$\pm$184.95 & 3362.58 & 0.4497 \\
jd400\_1 & 134302.39 & 137821.08$\pm$1868.47 & 10804.12 &  &  & 143045.76 & 146142.97$\pm$1767.66 & 10822.74 &  &  & 114537.34 & 116518.79$\pm$1422.92 & 10492.89 &  &  & \cellcolor[HTML]{D9D9D9}114489.28 & 115579.17$\pm$715.05 & 10474.83 & 0.1306 \\
jd400\_2 & 134084.89 & 135903.71$\pm$1220.10 & 11032.40 &  &  & 138520.94 & 141625.25$\pm$1845.82 & 10813.94 &  &  & 115303.21 & 116803.18$\pm$612.69 & 10409.36 &  &  & \cellcolor[HTML]{D9D9D9}114668.85 & \textbf{115658.43$\pm$606.08} & 10815.96 & 0.0015 \\
jd400\_3 & 115586.81 & 116862.96$\pm$798.97 & 10434.63 &  &  & 120065.47 & 123138.26$\pm$1663.56 & 10734.00 &  &  & \cellcolor[HTML]{D9D9D9}106951.59 & 107842.05$\pm$588.35 & 10547.36 &  &  & 107600.70 & 107927.90$\pm$351.95 & 10063.53 & 0.8206 \\
jd400\_4 & 135410.22 & 137420.59$\pm$1633.07 & 10744.00 & \multirow{-4}{*}{12.09\%} &  & 137863.84 & 141825.04$\pm$1917.25 & 10813.89 & \multirow{-4}{*}{16.49\%} &  & \cellcolor[HTML]{D9D9D9}126613.16 & 127573.34$\pm$914.21 & 10687.95 & \multirow{-4}{*}{-0.06\%} &  & 126927.71 & 127814.74$\pm$492.32 & 10362.35 & 0.2265 \\
jd600\_1 & 202030.04 & 207678.74$\pm$4499.67 & 18498.28 &  &  & 210658.62 & 215812.21$\pm$2820.95 & 17241.05 &  &  & 166524.71 & 168729.52$\pm$1110.96 & 14626.36 &  &  & \cellcolor[HTML]{D9D9D9}166113.91 & \textbf{166952.86$\pm$681.25} & 18016.97 & 0.0019 \\
jd600\_2 & 201625.22 & 205362.93$\pm$2455.99 & 18544.71 &  &  & 205943.56 & 211244.27$\pm$3406.55 & 16875.98 &  &  & 171996.76 & 173256.93$\pm$1387.16 & 14450.07 &  &  & \cellcolor[HTML]{D9D9D9}170362.24 & \textbf{171446.42$\pm$618.81} & 17416.13 & 0.0007 \\
jd600\_3 & 169213.04 & 171977.52$\pm$2543.89 & 18403.92 &  &  & 173106.97 & 176575.91$\pm$2006.10 & 15701.68 &  &  & 151879.40 & 153745.61$\pm$1118.40 & 16444.28 &  &  & \cellcolor[HTML]{D9D9D9}151185.33 & \textbf{152554.96$\pm$942.57} & 17806.93 & 0.0494 \\
jd600\_4 & 188018.90 & 191041.29$\pm$2450.90 & 18016.40 & \multirow{-4}{*}{19.58\%} &  & 197929.76 & 201306.90$\pm$2408.88 & 18013.41 & \multirow{-4}{*}{23.82\%} &  & 149893.97 & 151118.39$\pm$658.62 & 13495.48 & \multirow{-4}{*}{0.61\%} &  & \cellcolor[HTML]{D9D9D9}148729.43 & \textbf{149447.53$\pm$563.44} & 17432.51 & 0.0004 \\
jd800\_1 & 264947.80 & 271723.12$\pm$8551.65 & 25368.63 &  &  & 270951.52 & 280903.46$\pm$9941.59 & 23268.07 &  &  & 216194.15 & 217442.46$\pm$805.04 & 20030.45 &  &  & \cellcolor[HTML]{D9D9D9}213065.47 & \textbf{214228.94$\pm$974.71} & 25217.34 & 0.0002 \\
jd800\_2 & 263423.50 & 269372.58$\pm$3767.75 & 25209.97 &  &  & 268047.17 & 278252.06$\pm$12695.67 & 25214.92 &  &  & 226218.76 & 227630.94$\pm$1710.11 & 19306.58 &  &  & \cellcolor[HTML]{D9D9D9}222333.51 & \textbf{223685.24$\pm$1338.04} & 25214.00 & 0.0007 \\
jd800\_3 & 226143.04 & 232084.76$\pm$2892.31 & 26087.54 &  &  & 232052.94 & 244010.43$\pm$15964.86 & 21723.70 &  &  & 205429.64 & 207285.21$\pm$1202.21 & 23965.29 &  &  & \cellcolor[HTML]{D9D9D9}203751.04 & \textbf{205483.36$\pm$1355.49} & 24410.31 & 0.0082 \\
jd800\_4 & 248967.19 & 260163.60$\pm$9966.13 & 25228.11 & \multirow{-4}{*}{20.73\%} &  & 260642.67 & 267303.51$\pm$7378.45 & 25225.15 & \multirow{-4}{*}{24.19\%} &  & 194738.56 & 196802.01$\pm$952.89 & 23059.46 & \multirow{-4}{*}{1.26\%} &  & \cellcolor[HTML]{D9D9D9}192848.13 & \textbf{194144.22$\pm$987.70} & 24949.64 & 0.0005 \\
jd1000\_1 & 323977.19 & 337199.82$\pm$10774.46 & 32435.92 &  &  & 325780.40 & 329586.90$\pm$2529.39 & 32469.22 &  &  & 260502.17 & 262307.88$\pm$1199.71 & 29046.07 &  &  & \cellcolor[HTML]{D9D9D9}256626.16 & \textbf{257932.03$\pm$1114.15} & 32159.10 & 0.0002 \\
jd1000\_2 & 332121.12 & 341578.59$\pm$8647.74 & 32909.71 &  &  & 332438.38 & 335741.59$\pm$2112.72 & 32477.35 &  &  & 275972.37 & 278824.43$\pm$1564.13 & 28589.01 &  &  & \cellcolor[HTML]{D9D9D9}270788.78 & \textbf{273195.12$\pm$1537.57} & 32421.83 & 0.0002 \\
jd1000\_3 & 283186.67 & 288387.04$\pm$2503.00 & 33144.56 &  &  & 286423.55 & 290715.99$\pm$2296.68 & 32463.47 &  &  & 256229.79 & 257553.76$\pm$1262.58 & 25397.05 &  &  & \cellcolor[HTML]{D9D9D9}250802.03 & \textbf{253040.02$\pm$1411.33} & 31452.40 & 0.0002 \\
jd1000\_4 & 316816.36 & 323327.80$\pm$5526.63 & 32451.29 & \multirow{-4}{*}{23.65\%} &  & 320716.02 & 324125.48$\pm$3133.73 & 32407.54 & \multirow{-4}{*}{24.59\%} &  & 242662.53 & 244305.00$\pm$1250.99 & 27539.25 & \multirow{-4}{*}{1.83\%} &  & \cellcolor[HTML]{D9D9D9}238562.78 & \textbf{239966.65$\pm$1244.47} & 32276.48 & 0.0002 \\ 
\bottomrule
\end{tabular}%
}
\end{table*}

 \subsection{Results on Large-scale Instances}
To demonstrate the effectiveness of HMA for solving large-scale instances in the \( jd \) set, we compared HMA with three algorithms: Adapt-CMSA-STD~\cite{akbay2022application}, Adapt-CMSA-SETCOV~\cite{akbay2023cmsa}, and HMA without the BCD decomposition (denoted HMA w/o BCD).
% Here, we manually re-implemented Adapt-CMSA-STD and Adapt-CMSA-SETCOV because they are not open-sourced.
Since the original implementations of Adapt-CMSA-STD and Adapt-CMSA-SETCOV are not made open-source, we implemented them based on the descriptions provided in the literature, ensuring high fidelity to the reported algorithms.
% Here, we omit the existing algorithms~\cite{akbay2022application,akbay2023cmsa} for solving EVRP-TW-SPD because they are not open-sourced.
The BCD decomposition strategy integrated into HMA is tailored for large-scale problem instances as follows.
Specifically, instances with 200, 400, 600, 800, and 1000 customers were decomposed into 2, 4, 6, 8, and 10 subproblems, respectively, during the MS of HMA (see Algorithm~\ref{alg: Hybrid Search Framework}).
To evaluate the impact of runtime on algorithm performance for large-scale instances, both short and long time limits were imposed, i.e., the short time limits were set to 1800, 5400, 9000, 12600, and 16200 seconds (see Table~\ref{tab: large-scale short-time-limit}), while the long time limits were 3600, 10800, 18000, 25200, and 32400 seconds (see Table~\ref{tab: large-scale long-time-limit}). 
% Note that BCD requires the Cartesian coordinate of node $n_i$, while the \( jd \) set contains longitude and latitude. 
% Since the area covered is not large, we can use Mercator Projection to convert $(lng_i, lat_i)$ into $(x_i, y_i)$, specifically, $x_i = R \cdot \frac{\pi \cdot lng_i}{180}$, $y_i = R \cdot \ln \left( \tan \left( \frac{\pi}{4} + \frac{\pi \cdot lat_i}{360} \right) \right)$, where $R$ is the semi-major axis of the WGS84 ellipsoid model. 
Note that the BCD requires the Cartesian coordinates of node $n_i$, whereas the \( jd \) set contains the longitude and latitude values. 
Since the covered area is relatively small, we can use the Mercator projection to convert the geographic coordinates $(lng_i, lat_i)$ into Cartesian coordinates $(x_i, y_i)$. Specifically, the conversion is given by
$x_i = R \cdot \frac{\pi \cdot lng_i}{180}$, $y_i = R \cdot \ln \left( \tan \left( \frac{\pi}{4} + \frac{\pi \cdot lat_i}{360} \right) \right)$, 
where $R$ is the semi-major axis of the WGS84 ellipsoid model.
Tables~\ref{tab: large-scale short-time-limit} and~\ref{tab: large-scale long-time-limit} present the testing results of the four algorithms on large-scale instances in the \( jd \) set.
Unlike the \( akb\) set, for the \(jd\) set there is no priority for minimizing either
the number of used EVs or $TD$, thus the ``m'' values are not reported, but the ``best'' values are still included.
The ``avg$\pm$std'' columns represent the average and standard deviation of \( TC \) for the 10 best solutions obtained from 10 independent runs. 
The performance gap “$\Delta\%$” is measured as \( [TC_1 - TC_2]/{TC}_2 \), where \( TC_1 \) and \( TC_2 \) represent the best-found solutions from the other three competitors and HMA over 10 independent runs.
Here we show the average ``$\Delta\%$'' values for each problem size.
The best performance with respect to ``best'' is highlighted in gray. 
The following analysis focuses on two comparisons: 
i) the two HMA variants (i.e., HMA and HMA w/o BCD) against the two Adapt-CMSA variants (i.e., Adapt-CMSA-STD and Adapt-CMSA-SETCOV), and 
ii) HMA against HMA w/o BCD.

\subsubsection{HMA Variants versus Adapt-CMSA Variants} 
The results in Tables~\ref{tab: large-scale short-time-limit} and~\ref{tab: large-scale long-time-limit} clearly show that the two HMA variants consistently achieved superior performance over the two Adapt-CMSA variants in terms of both the best solution found and the average solution quality across 10 independent runs. 
A closer examination of the standard deviations for the two Adapt-CMSA variants reveals their unstable performance on large-scale instances, indicating they may struggle to escape poor-quality local optima. 
In contrast, the two HMA variants exhibit smaller standard deviations across all instances, suggesting a more reliable search process and stability.
The performance gap of the two Adapt-CMSA variants can be observed from the ``$\Delta\%$'' columns.
As the problem size increases, this ``$\Delta\%$'' value also increases under both short and long time limits, highlighting HMA's remarkable advantage in larger-scale instances.
This trend implies that, although Adapt-CMSA-STD and Adapt-CMSA-SETCOV may be adequate for smaller problems, they do not scale as effectively as HMA when confronted with larger instances. 

The scalability of HMA can be attributed to the following components. 
First, when tackling large-scale instances, the large neighborhood search of the HMA framework plays a crucial role, as it enables efficient exploration of the solution space. 
Second, the aggressive local search is indeed effective. 
From the analysis of solutions obtained by HMA, we observe that the utilization rate of charging stations is relatively low, and each route typically serves a small number of customers. 
This characteristic aligns well with practical scenarios.
It ensures that ALS can improve solutions in the VRP-TW-SPD solution space without being prematurely terminated due to the early occurrence of PSSI’s failure attempt (see Algorithm~\ref{algline: CDNS UNTIL2}, line~\ref{algline: CDNS UNTIL1}, the second termination condition). 
Moreover, this characteristic guarantees that the binary sequences initialized in PSI do not result in many infeasible solutions (see the ablation study on the \(jd\) set in Section~\ref{subsec: Effectiveness of Each Component in HMA}).

\subsubsection{HMA versus HMA w/o BCD} We further evaluated the impact of BCD by comparing HMA w/o BCD and HMA using the Wilcoxon rank-sum test at a 0.05 significance level (see the last column in Tables~\ref{tab: large-scale short-time-limit} and \ref{tab: large-scale long-time-limit}).
This comparison was necessary because the performance difference between the two HMA variants was not obvious (both average “$\Delta\%$” values on all instances are less than 1.00\% under different time limits).
The best performance with respect to ``avg$\pm$std'' is highlighted in bold, provided it is statistically significantly better than the other algorithm.
From Table~\ref{tab: large-scale short-time-limit}, it can be observed that under a short time limit, HMA consistently found better solutions on all instances of the \( jd \) set compared to HMA w/o BCD. 
The average performance of HMA is significantly better as the problem scale increases. 
In Table~\ref{tab: large-scale long-time-limit}, under a long time limit, HMA w/o BCD performed reversely better on some instances with 200 and 400 customers. 
This is due to the challenge of designing an ideal decomposition, which would partition all customers belonging to the same route in the optimal solution into the same subproblem. 
However, other decompositions, such as BCD, solve subproblems efficiently but may result in suboptimal solutions. 
Nonetheless, as the problem scale grows, the decomposition strategy still shows an obvious advantage on instances of larger scale with 600, 800, and 1000 customers since HMA obtained significantly higher-quality solutions. 
The performance gap ``$\Delta\%$'' between HMA w/o BCD and HMA across 10 independent runs is 0.61\%, 1.26\%, and 1.83\% averaged on instances with 600, 800, and 1000 customers, respectively. 
As the problem scale grows, this performance gap becomes even larger, confirming the effectiveness of the BCD decomposition in helping HMA solve very large EVRP-TW-SPD instances.

Overall, these experiments confirm that 
i) the proposed HMA framework provides the best and robust performance across all large-scale instances in the 
\(jd\) set, and 
ii) the BCD decomposition is particularly beneficial when tackling instances with 600 or more customers. 
Even with extended runtimes, HMA consistently outperforms the two Adapt-CMSA variants and yields significantly better solutions than its non-BCD counterpart on the instances with 400 or more customers~(see Section~\ref{subsec: statistical test} for the statistical test). 
Consequently, the scalability of HMA makes it well-suited for real-world EVRP-TW-SPD applications in which problem instances can easily grow to several hundred or even a thousand customers.

\begin{figure*}[t]
\centering
\captionsetup[subfigure]{labelformat=parens}
\setkeys{Gin}{width=\linewidth}
  \begin{minipage}[t]{0.32\linewidth}
    \centering
    \begin{subfigure}[b]{\linewidth}
      \includegraphics{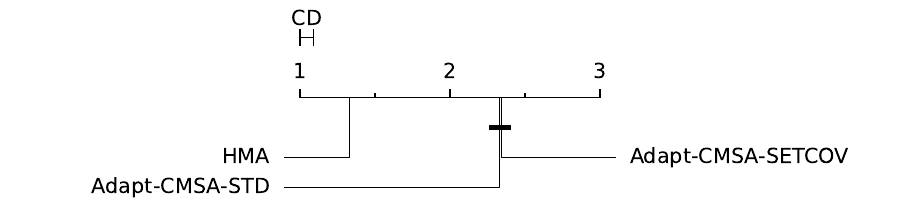}
      \caption{all instances}
      \label{fig:sub01}
    \end{subfigure}
    \vfill
    \begin{subfigure}[b]{\linewidth}
      \includegraphics{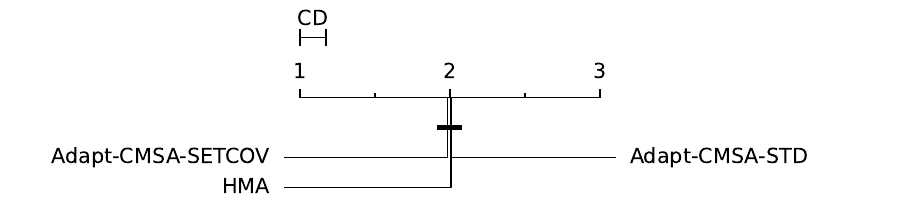}
      \caption{small-scale instances}
      \label{fig:sub02}
    \end{subfigure}
    \vfill
    \begin{subfigure}[b]{\linewidth}
      \includegraphics{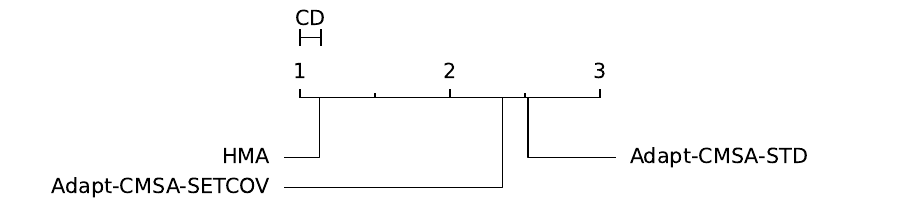}
      \caption{medium-scale instances}
      \label{fig:sub03}
    \end{subfigure}
    \vfill
    \begin{subfigure}[b]{\linewidth}
      \includegraphics{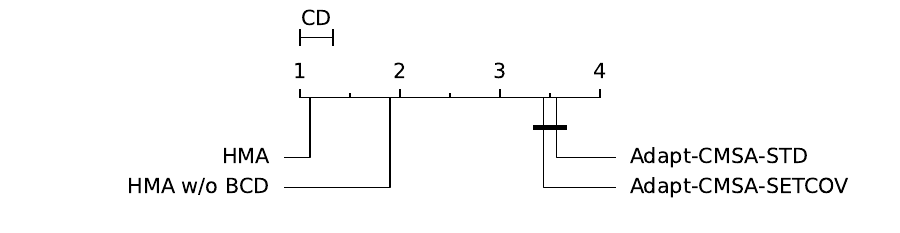}
      \caption{large-scale instances, a short time limit}
      \label{fig:sub04}
    \end{subfigure}
    \vfill
    \begin{subfigure}[b]{\linewidth}
      \includegraphics{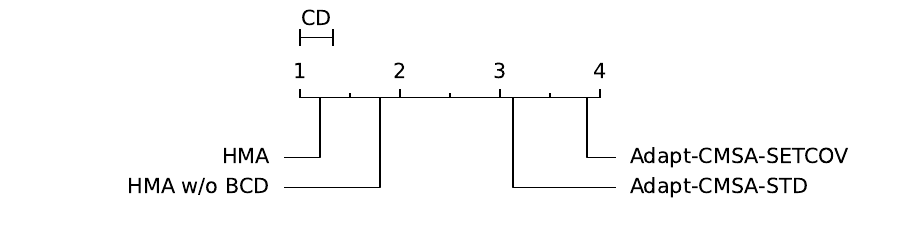}
      \caption{large-scale instances, a long time limit}
      \label{fig:sub05}
    \end{subfigure}
  \end{minipage}
  \hfill
  \begin{minipage}[t]{0.32\linewidth}
    \centering
    \begin{subfigure}[b]{\linewidth}
      \includegraphics{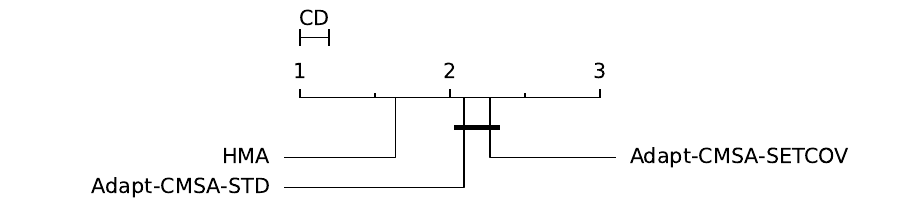}
      \caption{clustered instances}
      \label{fig:sub06}
    \end{subfigure}
    \vfill
    \begin{subfigure}[b]{\linewidth}
      \includegraphics{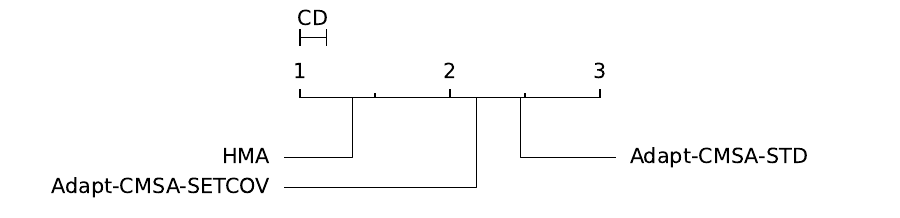}
      \caption{random instances}
      \label{fig:sub07}
    \end{subfigure}
    \vfill
    \begin{subfigure}[b]{\linewidth}
      \includegraphics{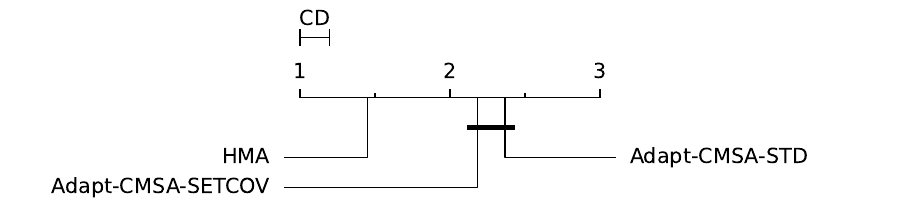}
      \caption{random-clustered instances}
      \label{fig:sub08}
    \end{subfigure}
    \vfill
    \begin{subfigure}[b]{\linewidth}
      \includegraphics{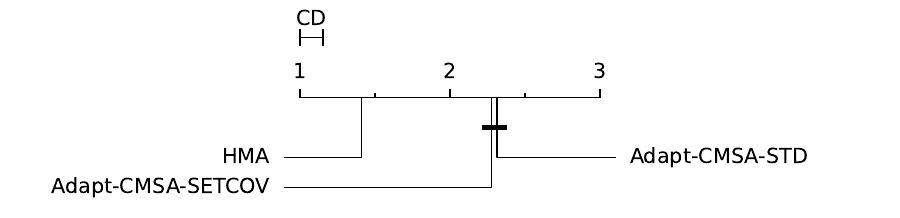}
      \caption{type-1 instances}
      \label{fig:sub09}
    \end{subfigure}
    \vfill
    \begin{subfigure}[b]{\linewidth}
      \includegraphics{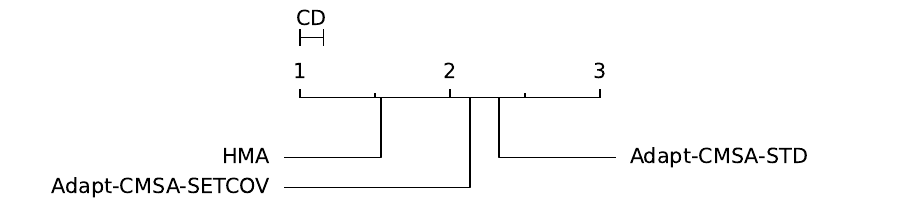}
      \caption{type-2 instances}
      \label{fig:sub10}
    \end{subfigure}
  \end{minipage}
  \hfill
  \begin{minipage}[t]{0.32\linewidth}
    \centering
    \begin{subfigure}[b]{\linewidth}
      \includegraphics{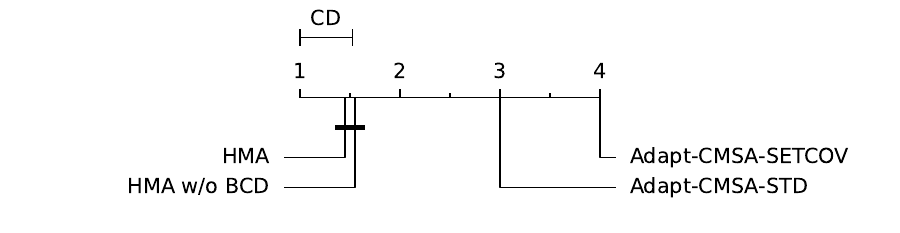}
      \caption{200-customer instances}
      \label{fig:sub11}
    \end{subfigure}
    \vfill
    \begin{subfigure}[b]{\linewidth}
      \includegraphics{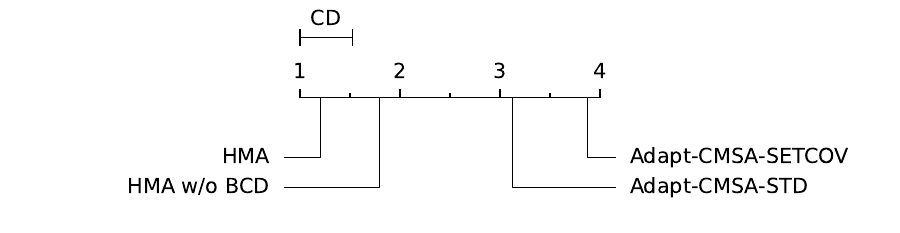}
      \caption{400-customer instances}
      \label{fig:sub12}
    \end{subfigure}
    \vfill
    \begin{subfigure}[b]{\linewidth}
      \includegraphics{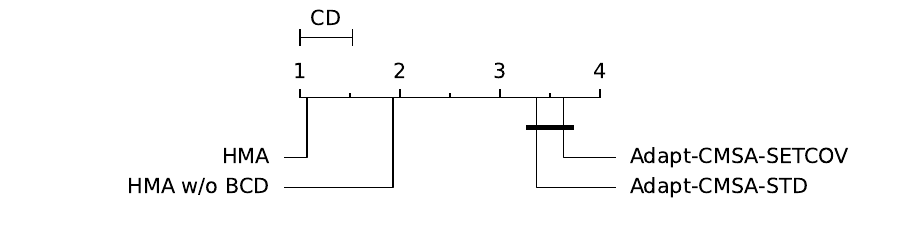}
      \caption{600-customer instances}
      \label{fig:sub13}
    \end{subfigure}
    \vfill
    \begin{subfigure}[b]{\linewidth}
      \includegraphics{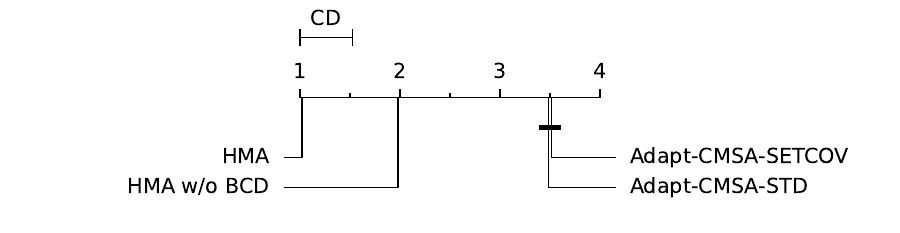}
      \caption{800-customer instances}
      \label{fig:sub14}
    \end{subfigure}
    \vfill
    \begin{subfigure}[b]{\linewidth}
      \includegraphics{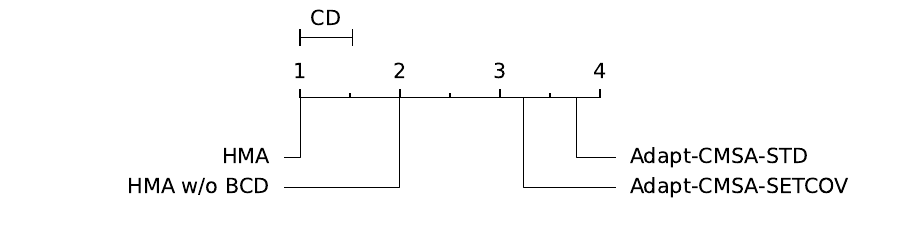}
      \caption{1000-customer instances}
      \label{fig:sub15}
    \end{subfigure}
  \end{minipage}
\caption{Critical difference (CD) plots across various problem instances using the Nemenyi post hoc test.}
\label{fig:all cd plots}
\end{figure*}

\subsection{Statistical Test of Algorithm Performance and Robustness}
\label{subsec: statistical test}
To support our claim that HMA has a significant performance advantage over existing methods across a wide range of problem instances, we also present critical difference (CD) plots. 
These plots serve as a statistical tool to help evaluate the obtained results in Tables~\ref{tab: small-scale}, \ref{tab: medium-scale}, \ref{tab: large-scale short-time-limit} and \ref{tab: large-scale long-time-limit}, and to measure the algorithm's robustness when comparing multiple algorithms over various problem instances~\cite{akbay2023cmsa}.

 The procedure of the statistical test is described as follows. 
 We first apply the Friedman test to simultaneously compare all algorithms and determine if significant differences exist in their rankings. 
 If the null hypothesis, which assumes equal performance among the algorithms, is rejected, we proceed with pairwise comparisons using the Nemenyi post hoc test. 
 The results are then graphically displayed in CD plots (see Figure~\ref{fig:all cd plots}), where each algorithm is positioned along the horizontal axis according to its average rank over the specific subset of problem instances. 
In total, we consider fifteen subsets. 
Sub-figures~\ref{fig:sub01}--\ref{fig:sub05} illustrate various problem scales, Sub-figures~\ref{fig:sub06}--\ref{fig:sub10} show different customer distributions and constraint characteristics (see Section~\ref{benchmark}), and Sub-figures~\ref{fig:sub11}--\ref{fig:sub15} provide a more in-depth analysis of large-scale scenarios.
Algorithms that exhibit performance differences within the critical difference threshold, computed at a significance level of 0.05, are considered statistically equivalent, as indicated by horizontal lines connecting their markers.

 As shown in Figure~\ref{fig:all cd plots}, HMA is statistically significantly better than its competitors in almost all subsets, except for the small-scale instances. 
 In this subset, Adapt-CMSA-STD, Adapt-CMSA-SETCOV, and HMA show comparable performance, with no significant differences. 
 In Sub-figures~\ref{fig:sub01}--\ref{fig:sub05}, Adapt-CMSA-SETCOV outperforms Adapt-CMSA-STD on medium-scale instances; however, in large-scale instances with a long time limit, the results are reversed. 
 HMA performs better than HMA w/o BCD in large-scale instances under different time limits. 
 In Sub-figures~\ref{fig:sub06}--\ref{fig:sub10}, HMA demonstrates greater strength on random, random-clustered, and type-1 instances. Adapt-CMSA-SETCOV is the runner-up on \( akb \) set; 
 In Sub-figures~\ref{fig:sub11}--\ref{fig:sub15}, HMA and HMA w/o BCD perform similarly in instances with 200 customers, but a significant difference emerges as the scale increases from 400 to 1000 customers. Adapt-CMSA-STD performs better than Adapt-CMSA-SETCOV with 200 and 400 customers; however, in instances with 1000 customers, the results are reversed. 
 Overall, these CD plots demonstrate that HMA exhibits the best and the most robust performance across different instance scales, customer distributions, constraint characteristics, and time limits.

\subsection{Effectiveness of PSSI, CDNS, and Hybrid Framework}
\label{subsec: Effectiveness of Each Component in HMA}
\begin{figure}[h]
    \centering
    \includegraphics[width=1\linewidth]{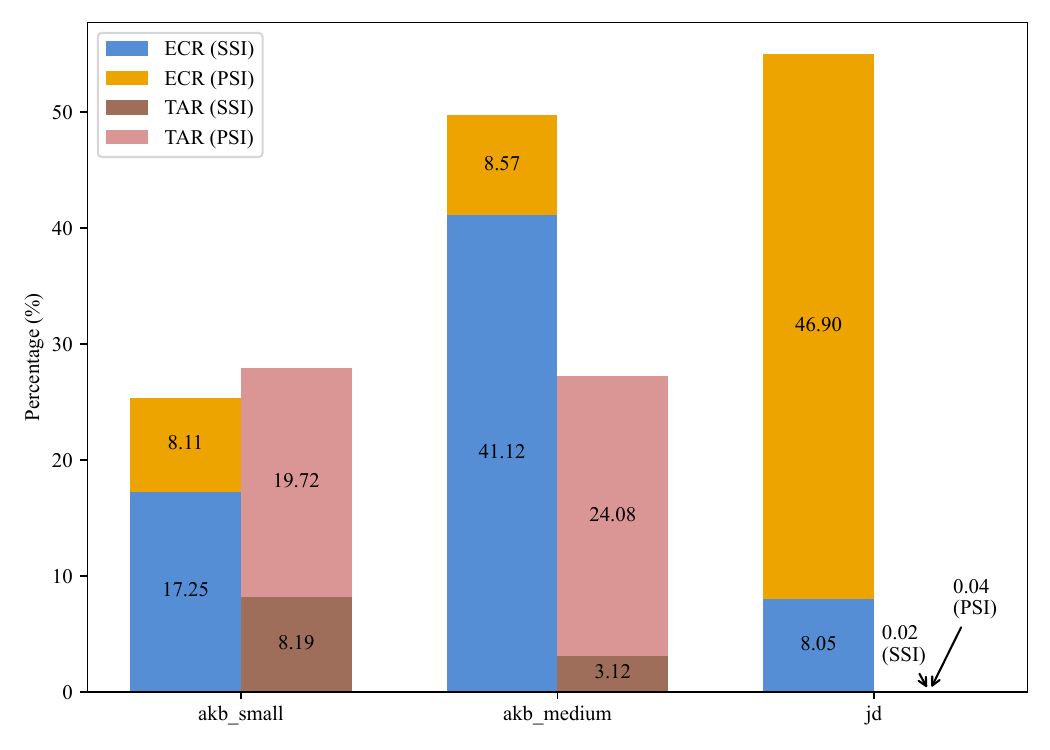}
    \caption{Comparative effects of PSI and SSI on each subset. The left stacked bar illustrates the exclusive contribution ratio (ECR) of SSI and PSI, with the remaining portion corresponding to shared contributions. The right stacked bar depicts the time allocation ratio (TAR) of SSI and PSI, with the remainder representing the TAR of other algorithm components.}
    \label{fig: SSI vs PSI}
\end{figure}

First, we separately analyzed the comparative effects of PSI and SSI for the following reasons: 
PSSI selects the best result from PSI and SSI.
Therefore, we are more concerned about which strategy, PSI or SSI, achieves better station insertion on a route. 
% Note that SSI produces only a single route, compared to PSI with a population. 
Note that unlike PSI, which maintains a population of routes, SSI produces only a single route.
We need to validate whether PSI can indeed find other higher-quality station insertion configurations with acceptable additional computational cost.
To this end, we ran HMA once on all instances, tracking the performance of PSI and SSI in terms of station insertion on the same route. 
Figure~\ref{fig: SSI vs PSI} illustrates this performance. The exclusive contribution ratio (ECR) is calculated as $n_1/n_2$ for each instance, where $n_1$ is the number of insertion results obtained exclusively from SSI or PSI and $n_2$ is the total number of insertion results.
The time allocation ratio (TAR) is computed as $t_1/t_2$ for each instance, where $t_1$ is the cumulative time consumed in SSI or PSI components, and $t_2$ is the total time consumed in HMA. 
Note that the left bar labeled ``ECR'' and the right bar labeled ``TAR'' in Figure~\ref{fig: SSI vs PSI} represent the average values for each subset. 

The results clearly indicate that PSI can indeed find other higher-quality station insertions, as an ECR greater than 0 shows that it found some insertions that are strictly better than those produced by SSI.
% In the \(jd\) set, PSI achieved an ECR of 46.90\%, which represents a significant contribution to PSSI. 
% In the $jd$ set, the ECR of PSI even achieved 46.90\%, accounting for a large proportion of contributions to PSSI.
In the \(jd \) set, PSI achieved an ECR of 46.90\%, accounting for a substantial portion of PSSI's overall performance.
Notably, although PSI with a genetic algorithm required approximately $2$--$8$ times the runtime of SSI, this additional time is worthwhile and acceptable, since SSI cannot further improve its insertion results.
% Notably, although PSI with a genetic algorithm took approximately $2\sim8$ times runtime than SSI, this additional runtime is worthwhile and acceptable since SSI cannot further improve its insertion results.
Thus, combining the results of both PSI and SSI can better avoid local optima and construct higher-quality routes with acceptable additional computational cost.

Second, we conducted an ablation study to evaluate the effectiveness of other components in HMA.
We tested the following four HMA variants on the $akb$ and $jd$ sets, each of which is different from HMA in one component:
\begin{enumerate}
    \item w/o CLS: lines~\ref{algline: CDNS REPEAT2}--\ref{algline: CDNS UNTIL2} of Algorithm~\ref{alg: Cross-Domain Neighborhood Search} are removed;
    \item w/o ALS: lines~\ref{algline: CDNS x copy}--\ref{algline: CDNS UNTIL1} of Algorithm~\ref{alg: Cross-Domain Neighborhood Search} are removed;
    \item w/o MS: lines~\ref{algline: HS population introduction}--\ref{algline: HS until improvement} of Algorithm~\ref{alg: Hybrid Search Framework} are removed;
    \item w/o LNS: lines~\ref{algline: HS while}--\ref{algline: HS if check}, \ref{algline: HS end if}--\ref{algline: HS endwhile} of Algorithm~\ref{alg: Hybrid Search Framework} are removed.
\end{enumerate}
The first two variants were to validate the effectiveness of CDNS, while the latter two were to validate the effectiveness of the hybrid framework.
For each variant, the performance degradation ratio (PDR) is computed as $[TC_1 - TC_2]/{TC}_2$ for each instance, where $TC_1$ and $TC_2$ are the average $TC$ obtained by the variant and HMA across 10 independent runs, respectively. 
A PDR greater than 0 indicates the component is useful for this
specific instance. 
Larger PDR values indicate a greater impact. 
Table~\ref{tab: HMA variants} presents the average PDR on each subset.

\begin{table}[h]
\caption{Average PDR for the four HMA variants on each subset.}
\label{tab: HMA variants}
\resizebox{\columnwidth}{!}{%
\begin{tabular}{ccccc}
\toprule
Subset & w/o CLS & w/o ALS & w/o MS & w/o LNS \\ \midrule
$akb_{small}$ & 1.80\% & 0.00\% & \cellcolor[HTML]{D9D9D9}2.94\% & 0.00\% \\
$akb_{medium}$ & 1.03\% & 0.06\% & \cellcolor[HTML]{D9D9D9}3.49\% & 0.37\% \\
$jd$ & 0.00\% & \cellcolor[HTML]{D9D9D9}4.03\% & 0.60\% & 2.47\% \\ \midrule
Avg.PDR & 0.84\% & 0.77\% & \cellcolor[HTML]{D9D9D9}2.98\% & 0.75\% \\ \bottomrule
\end{tabular}%
}
\end{table}

The results show that, on average, no component negatively affects the performance of HMA.
Specifically, the ``w/o CLS'' and ``w/o ALS'' results show that CLS benefits small-scale and medium-scale instances, while ALS is more effective for large-scale instances. 
CLS is good at refined search, which is beneficial for small solution spaces. 
In contrast, ALS leverages existing efficient VRP-TW-SPD move evaluation, advantageous for large solution spaces that require fast exploration.
Similarly, the ``w/o MS'' and ``w/o LNS'' results indicate that MS is more effective for small-scale and medium-scale instances due to population diversity, whereas a fast search of LNS benefits large-scale instances. 
The population maintains solution diversity in smaller problems; however, as the problem scale increases, the dominant role of individual-based search capabilities becomes more evident.

\section{Conclusion}
\label{sec: Conclusion}
This work investigated an emerging routing problem, namely EVRP-TW-SPD, and proposed a hybrid memetic algorithm called HMA to solve it.
HMA integrates two novel components: PSSI for handling partial recharges and CDNS as the local search procedure.
Both components can be easily applied to other EVRP variants.
% The population-based search is most beneficial for small-scale and medium-scale instances, while aggressive local search matters in large-scale instances.
Moreover, a new, large-scale benchmark set was established from a real-world JD logistics application.
This set contains 20 problem instances with 200, 400, 600, 800, and 1000 customers, each with 100 stations. 
Comprehensive experiments showed that HMA outperformed existing methods across a wide range of instances.

Although HMA has shown excellent performance in the experiments, several promising directions exist for further enhancing HMA's performance.
First, the PSSI procedure could benefit from a learned function (e.g., as a neural network~\cite{LiuZTY23}) that evaluates insertions instead of heuristic rules, which might lead to higher-quality feasible routes within even smaller time budget.
Second, for large-scale problems, dividing the original problem into smaller, manageable subproblems enhances efficiency and yields high-quality solutions.
However, the success of this approach heavily depends on the effectiveness of the decomposition strategy.
Rather than relying on hand-designed strategies, future improvements could leverage learning-based methods~\cite{tang2024learn} or general-purpose pretrained models~\cite{LiuCQ0O24} to automatically design effective decomposition strategies tailored to specific customer distributions or constraint characteristics.

\begin{appendices}
\section{Proof of Proposition 1}
\label{app:proof_pro_1}

\begin{proof}
Consider a feasible EVRP-TW-SPD solution $\bm{S}$. By definition, $\bm{S}$ satisfies all electricity, capacity, and time window constraints.
Let $\bar{\bm{S}}$ be the solution obtained by removing all charging stations in $\bm{S}$.
Suppose the travel time between any two nodes satisfies the triangle inequality; then, for any two successive customers $c_i$ and $c_j$ in $\bar{\bm{S}}$ that are separated by a charging station $f$, the triangle inequality ensures:
\[
t_{c_ic_j} \leq t_{c_if} + t_{fc_j} + g \cdot q_f,
\]
where $g \cdot q_f$ is the charging time at station $f$.
This inequality implies that the travel time between customers in $\bar{\bm{S}}$ is no greater than in $\bm{S}$.
Consequently, $\bar{\bm{S}}$ satisfies all capacity and time window constraints of VRP-TW-SPD, making it a feasible solution.

In cases where the travel time does not satisfy the triangle inequality, we can preprocess the problem instance to ensure this property holds.
Specifically, we first apply the Floyd-Warshall algorithm to compute the shortest path in terms of travel time between every pair of nodes.
These shortest paths are allowed to pass through charging stations but not through the depot or other customers.
We then reconstruct the edge set $\bm{E}$ using these shortest paths, defining each new edge (or ``hyperarc'') as $(i,j)$ to be the shortest path from node $i$ to $j$.
The travel time associated with these new edges naturally satisfies the triangle inequality, allowing Proposition \ref{proposition: VRP-TW-SPD & EVRP-TW-SPD} to hold for any problem instance.
\end{proof}

\end{appendices}
% Biography text here without a photo.

% \begin{IEEEbiography}[{\includegraphics[width=1in,height=1.25in,clip,keepaspectratio]{fig1.png}}]{IEEE Publications Technology Team}
% In this paragraph you can place your educational, professional background and research and other interests.\end{IEEEbiography}

\bibliographystyle{IEEEtran}
\bibliography{IEEEabrv,references}	

\end{document}